\documentclass[11pt]{article}

\usepackage[preprint]{acl}

\usepackage{times}
\usepackage{latexsym}

\usepackage[T1]{fontenc}
\usepackage[utf8]{inputenc}

\usepackage{microtype}

\usepackage{inconsolata}

\usepackage{graphicx}

\usepackage{subcaption}
\usepackage{enumitem}
\usepackage{multirow}
\usepackage{nicefrac}
\usepackage{amsmath}
\usepackage{booktabs}
\usepackage{xcolor}
\usepackage{hyperref}

\usepackage{comment}
\usepackage{subcaption}
\usepackage{color-edits}
\addauthor[Emmy]{el}{purple}
\addauthor[Steven]{sf}{blue}
\addauthor[Varun]{vg}{orange}
\addauthor[Karan]{ks}{cyan}
\addauthor[Michael]{my}{yellow}
\addauthor[Derek]{dt}{violet}
\addauthor[Sachin]{sk}{red}

\newcommand\blfootnote[1]{%
  \begingroup
  \renewcommand\thefootnote{}\footnote{#1}%
  \addtocounter{footnote}{-1}%
  \endgroup
}

\newcommand{\loss}{\mathcal{L}}

\title{To Memorize or to Retrieve: Scaling the\\Interaction Between Pretraining and Retrieval}
\author{
  \textbf{Karan Singh\textsuperscript{1,\textdagger}},
  \textbf{Michael Yu\textsuperscript{2,\textdagger}},
  \textbf{Varun Gangal\textsuperscript{3,\textdagger}},
\\
  \textbf{Zhuofu Tao\textsuperscript{2,\textdagger}},
  \textbf{Sachin Kumar\textsuperscript{4,\textdagger}},
  \textbf{Emmy Liu\textsuperscript{5,\textdagger}},
  \textbf{Steven Y. Feng\textsuperscript{1,\textdagger}}
\\
\\
  \textsuperscript{1}Stanford University \quad
  \textsuperscript{2}Independent Researcher \quad
  \textsuperscript{3}Patronus AI
\\
  \textsuperscript{4}The Ohio State University \quad
  \textsuperscript{5}Carnegie Mellon University \quad
  \textsuperscript{\textdagger}DegenAI Labs
}

\begin{document}
\maketitle
\begin{abstract}
Retrieval-augmented generation (RAG) improves language model (LM) performance by providing relevant context at test time for knowledge-intensive situations. 
In this work, we systematically study the trade-off between pretraining and retrieval by training OLMo-2-based LMs ranging from 30M to 3B parameters on up to 100B DCLM tokens, while varying pretraining data scale, retrieval store size, and retrieval store source (pretraining vs. new data) across reasoning, scientific QA, and open-domain QA benchmarks.
We find that retrieval gains depend on model capacity and pretraining exposure and are strongly front-loaded, with a median 91\% of the largest observed improvement realized by one retrieval token per model parameter. However, the interaction is objective-dependent: smaller models gain more in gold-answer perplexity, whereas larger, more-pretrained models gain more in accuracy. Retrieval from previously seen data also preserves most of the held-out retrieval gain. Retrieval is therefore a task-, regime-, and metric-dependent complement to parametric learning whose value also depends on datastore size and information novelty. Overall, this motivates the explicit partitioning of data between internalization and external access for LM design.
\blfootnote{Code for our paper is available at \url{https://github.com/DegenAI-Labs/RAG-Scaling-Laws}. Correspondence to \href{mailto:karanps@stanford.edu}{karanps@stanford.edu}, \href{mailto:vgtomahawk@gmail.com}{vgtomahawk@gmail.com}, and \href{mailto:syfeng@stanford.edu}{syfeng@stanford.edu}.}
\end{abstract}

% We find that retrieval gains are not additively separable from the generator's training regime, and vary with model capacity and pretraining exposure. Crucially, the direction of this interaction depends on the evaluation target: likelihood gains are larger for smaller models, whereas accuracy gains are larger for larger, more-pretrained models. 
% Together, our results show that retrieval’s value depends not only on how much external data is available, but also on whether that information is novel and what the model has already learned, providing a quantitative account of when external access to knowledge complements or substitutes further pretraining. 

\begin{figure*}[t]
\begin{center}
\centerline{\includegraphics[width=\textwidth]{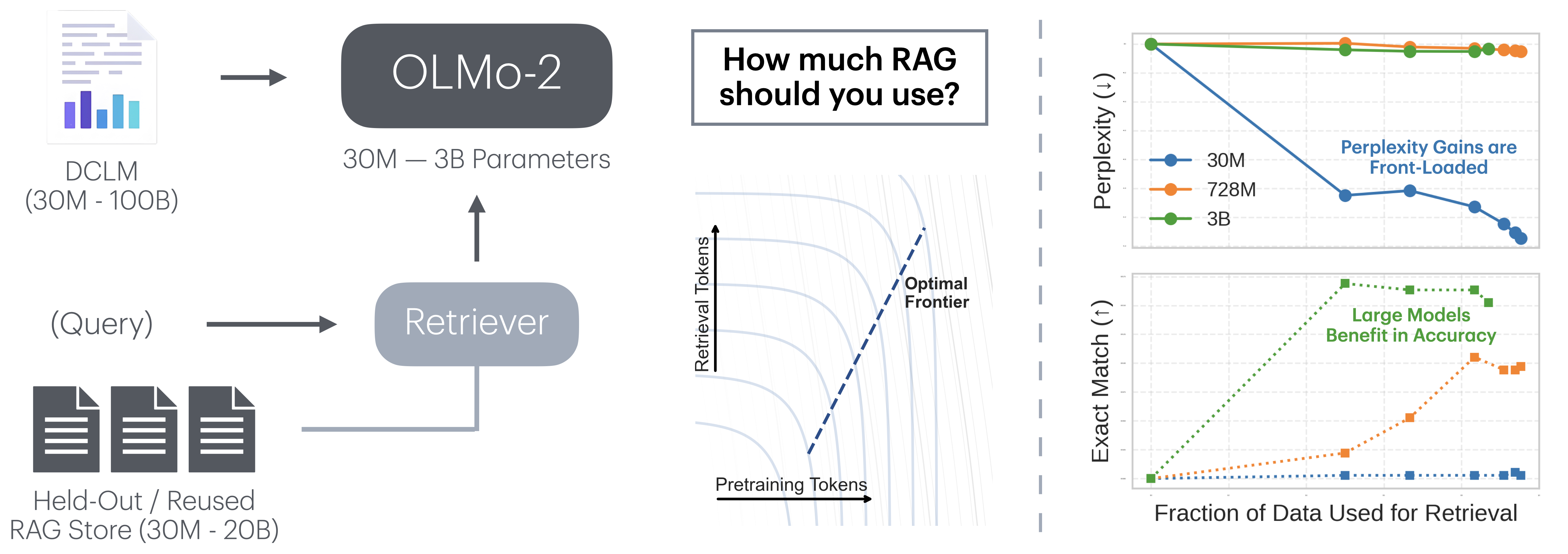}}
\caption{\textbf{Overview of the controlled pretraining--retrieval scaling study.}
\textit{Left:} We train OLMo-2 models from 30M to 3B parameters while varying DCLM pretraining exposure, datastore size, and whether indexed data is held out from or reused during pretraining.
\textit{Center:} We fit a scaling law over model size, pretraining tokens, and retrieval tokens; fixed-token-budget slices support performance-side comparisons among allocations.
\textit{Right:} The interaction with model scale depends on the evaluation target. Gold-answer perplexity gains are front-loaded and larger for smaller models, whereas decision-accuracy gains favor larger models on the multiple-choice benchmarks studied.}
\label{fig:intro}
\vskip -0.6cm
\end{center}
\end{figure*}

\section{Introduction}

Scaling laws \citep{Hestness2017DeepEmpirically,Kaplan2020ScalingModels} have established how language model (LM) performance improves with parameters and training tokens, but they treat the training corpus as monolithic. In standard pretraining, all available data is consumed parametrically, implicitly assuming that knowledge should be compressed into model weights. Retrieval-augmented generation (RAG) introduces a new degree of freedom: a portion of the corpus can instead be held out as an external datastore and accessed at inference time. These two uses of data %, parametric memorization versus non-parametric retrieval,
are fundamentally different, with distinct computational costs, inductive biases, and failure modes. For example, parametric learning may lead to an inaccurate internal world model and hallucination tendencies \citep{liu2026halluworldcontrolledbenchmarkhallucination,Liu2026AStupid}, while RAG may lead to errors from retrieving irrelevant or misleading documents %, causing the model to ground its generation in incorrect context
\citep{Lewis2021Retrieval-AugmentedTasks}. There are also ties to cognition: people typically internalize abstract reasoning skills while relying on external memory (e.g., books, search engines, notes) for factual recall \citep{Wegner1987,Risko2016CognitiveOffloading,Sparrow2011GoogleEffects,norman1988psychology,clark1998extended}. While we do not directly study these mechanisms, they motivate viewing parametric and non-parametric knowledge as complementary resources, raising the question of how the value of external access changes as the generator acquires more parametric knowledge.

We study the question of how to allocate data between parametric knowledge and external knowledge through controlled experiments. Concretely, we draw both pretraining and retrieval data from the same DCLM corpus. For each model size, we vary the number of tokens used for pretraining $D$ and placed in the retrieval datastore $R$. In our main condition, these sets are disjoint: increasing $R$ gives the model inference-time access to information it did not encounter during pretraining. We also construct datastores from the model's own pretraining data, testing whether retrieval remains useful when it provides easier access to previously seen information rather than new information. We fit a scaling law over model size, pretraining tokens, and datastore size, and use it to compare pretraining and retrieval within controlled token-budget slices.

Pretraining internalizes information in model parameters, while retrieval makes corpus information available as context at inference time.
Understanding the tradeoff between these two mechanisms is essential for designing efficient, scalable language model systems. We study this empirically across model scales ranging from 30M to 3B parameters, systematically varying the amount of pretraining data and the size of the retrieval datastore constructed from the same corpus (Figure~\ref{fig:intro}). We evaluate across a diverse set of benchmarks spanning multiple domains and knowledge types. In summary, we make the following contributions:

\begin{itemize}[topsep=0pt,leftmargin=6mm,itemsep=0pt,partopsep=0pt,parsep=2pt]
    \item We introduce an interaction scaling law and show that retrieval gains depend on model size, pretraining exposure, and evaluation target: smaller, less-pretrained models gain more in gold-answer likelihood, while larger, more-pretrained models gain more in choice-normalized probability and accuracy. 
    \item This distinction separates likelihood scaling, which measures support for the correct answer, from decision-level scaling, which is relevant when optimizing downstream accuracy.
    \item We show that retrieval gains are strongly front-loaded: a median \(91\%\) of the largest observed gain is realized with one datastore token per model parameter, followed by sharply diminishing improvements at larger datastore sizes.
    \item By comparing held-out and reused datastores, we separate the benefit of accessing novel information from that of re-accessing information seen during pretraining. Reuse preserves \(72\text{--}92\%\) of the held-out retrieval gain.

    % We show that the relationship between pretraining and retrieval is structured but non-trivial, with retrieval yielding scale- and regime-dependent, non-monotonic effects.

    % \item We identify a scale-dependent crossover point beyond which retrieval becomes an efficient substitute for pretraining. We demonstrate that retrieval can act as an efficient substitute for pretraining beyond a scale-dependent crossover point, before which RAG either hurts or does not affect performance, and after which additional retrieval yields increasing returns relative to further pretraining. 
    % At the same time, the marginal benefit of retrieval diminishes as models approach saturation, indicating that the optimal allocation depends critically on both model capacity and training regime. 
    % We further observe that the impact of retrieval varies across tasks, with knowledge-intensive benchmarks benefiting more than others.
\end{itemize}

\noindent Overall, we show that pretraining and retrieval cannot be analyzed independently: retrieval's value depends on model capacity, pretraining exposure, task, and whether the indexed information was previously seen. Our scaling framework provides a controlled account of when external access complements or substitutes for additional pretraining, motivating a shift in how pretraining corpora should be conceptualized and helping to inform LM design.

% Taken together, our findings establish a unified scaling perspective on parametric and non-parametric knowledge, and provide practical guidance for RAG-aware training. Rather than treating pretraining and retrieval as separate design choices, we show they can be jointly optimized under a fixed data budget, enabling efficient use of large-scale corpora.

\section{Related Works}
\subsection{Scaling Laws for Pretraining}

Many works study how LM performance scales with model size, dataset size, and compute.
\citet{Kaplan2020ScalingModels} established predictable power-law relationships between these factors. %, enabling extrapolation from smaller-scale experiments
Chinchilla later showed that compute-optimal training requires jointly scaling model and data size \citep{Hoffmann2022TrainingModels}.
More recent work extended this: % beyond the original Chinchilla regime:
\citet{Gadre2024LanguageTasks} show that scaling laws remain predictive in overtrained regimes and relate pretraining loss to downstream task performance, while other work incorporates data mixture \citep{Ye2025DataPerformance,Shukor2025ScalingMixtures} %, %data quality \citep{Subramanyam2026ScalingPretraining},
and domain-specific continual pretraining \citep{Que2024D-CPTModels}.

These works suggest that pretraining efficiency depends not only on parameters and tokens, but also on data and its proportions. A related line of work studies scaling when inference cost matters.
\citet{Sardana2024BeyondLaws} show that %Chinchilla-optimal training is not necessarily deployment-optimal, and that 
smaller models trained longer can be preferable when inference demand is high.
\citet{Bian2025ScalingModels} incorporate architecture-aware latency into scaling analysis, showing that parameter count alone is an incomplete proxy for efficiency. These results motivate treating pretraining, model size, data quality, and cost as coupled optimization problems.

\subsection{Retrieval-Augmented Language Models}

Retrieval-augmented LMs address a key limitation of purely parametric LMs: knowledge is stored implicitly in weights, making updates expensive and provenance difficult to trace. REALM was among the first to integrate retrieval directly into pretraining by jointly learning a dense retriever with a masked-LM objective \citep{Guu2020REALM:Pre-Training}.
RAG popularized retrieval-augmented generation for knowledge-intensive tasks by conditioning a generator on retrieved Wikipedia passages \citep{Lewis2021Retrieval-AugmentedTasks}, while kNN-LM showed that nearest-neighbor lookup %over a datastore of model activations 
can improve perplexity and domain adaptation \citep{Khandelwal2020GeneralizationModels}.

Subsequent work scaled this to larger corpora and general-purpose LMs. RETRO demonstrated that large external datastores can match larger parametric models \citep{Borgeaud2022ImprovingTokens}, and Atlas showed strong few-shot performance with retrieval-augmented models that can be updated independently of the generator \citep{Izacard2022Atlas:Models}. 
%Later extensions %broadened the design space further, including %black-box augmentation via retrieved context (REPLUG) \citep{Shi2023REPLUG:Models}
%include adding external memory mechanisms (Memorizing Transformers) \citep{Wu2022MemorizingTransformers} and adaptive retrieval with self-reflection (Self-RAG) \citep{Asai2023Self-RAG:Self-Reflection}.
Recent surveys emphasize trade-offs among retriever quality, memory freshness, grounding, and system complexity \citep{Hu2025RAGProcessing}. While much of the RAG literature focuses on improving downstream factuality at test time, systems such as REALM, RETRO, and Atlas suggest that retrieval can alter the pretraining trade-off itself by offloading knowledge from parameters into external memory. 

\subsection{Small Language Models: Pretraining, Evaluation \& Data Efficiency}

% Recent work on small language models (SLMs) emphasized that performance under tight parameter budgets depends heavily on architecture, data quality, and training duration.
% TinyLlama showed that a 1.1B model trained on $\sim$1T tokens can substantially outperform earlier open models of similar size \citep{Zhang2024TinyLlama:Model}.
% %MobileLLM found that deeper, thinner architectures with efficiency-oriented design choices improve quality in the sub-billion regime \citep{Liu2024MobileLLM:Cases}.
% SmolLM2 %pushed the data-centric perspective further,
% showed that a 1.7B model overtrained on a careful mixture of web, math, code, and instruction data can outperform several recent baselines \citep{Allal2025SmolLM2:Model}.
% %In parallel, Sheared LLaMA showed that structured pruning followed by continued pretraining can produce competitive small models without full training from scratch \citep{Xia2024ShearedPruning}.

% shortened
Recent work on small language models (SLMs) emphasizes that performance under tight parameter budgets depends heavily on architecture, data quality, and training duration. TinyLlama and SmolLM2 show that prolonged pretraining and careful data curation can make ${\sim}$1B-parameter models competitive with larger open baselines \citep{Zhang2024TinyLlama:Model,Allal2025SmolLM2:Model}.
At smaller data budgets, BabyLM studies learning from only 10M--100M tokens \citep{warstadt2023findings,hu2024findings}, motivating analyses of inductive biases and linguistic asymmetries \citep{kallini2024mission,hu2025production,feng2026babyscaleinvestigatingmodels,zeng2026bringingbilingualbabylminvestigating}.

Evaluation is especially consequential for SLMs. HELM advocates multi-metric, scenario-based evaluation \citep{Liang2023HolisticModels}, while DataComp-LM combines standardized pretraining recipes with broad downstream evaluation \citep{Li2025DataComp-LM:Models}. Data-centric studies further show that specialized corpora can strengthen compact models \citep{Gunasekar2023TextbooksNeed,Penedo2024TheScale,Allal2025SmolLM2:Model}, while concept, knowledge, and visual augmentation can improve commonsense generation \citep{lin-etal-2020-commongen,feng-etal-2021-sapphire,Feng_Lu_Tao_Alikhani_Mitamura_Hovy_Gangal_2022,feng-etal-2023-chard}. 

Under fixed training budgets, deduplication \citep{Lee2022DeduplicatingBetter}, educational filtering \citep{Penedo2024TheScale}, domain reweighting \citep{Xie2023DoReMi:Pretraining}, and mixture optimization \citep{Ye2025DataPerformance} can improve data efficiency. Related work studies data mixtures across pretraining stages \citep{Feng2024MaximizePretraining,liu2026midtrainingbridgespretrainingposttraining}, mixtures of code and target-domain data \citep{Ma2023AtReasoning,Baek2026TheData}, and data ordering or progressive model scaling curricula \citep{Feng2024IsModels,Singh2026Curriculum-GuidedPretraining}. Together, these works show that data efficiency depends on what, when, and how information is presented.

\subsection{Retrieval-Scaling}

Prior work establishes that retrieval gains persist across datastore, model, and inference-compute scales. \citet{Shao2024ScalingDatastore} jointly vary generator size, pretraining exposure, and datastore size, showing that retrieval can shift compute-optimal frontiers. \citet{Fang2025ReusingMultiplier} retrieve from models' own pretraining corpora and show that previously encountered data retains test-time value, while \citet{Yue2025InferenceGeneration,Zhou2026InferenceGeneration} characterize how performance changes with retrieval and generation compute at inference time. Together, these studies demonstrate that both new and previously seen information can improve pretrained models.

\paragraph{Summary and Motivation.}
Prior work shows that language-model performance scales with pretraining resources, datastore size, and inference-time computation, and that even previously encountered data can retain value when retrieved at test time. What remains unclear is how retrieval's marginal value changes as a model acquires more parametric knowledge, and how much of that value depends on whether the retrieved information was seen during pretraining. We address this gap through a controlled factorial study: we train matched OLMo-2 models from scratch across model and pretraining scales, vary datastore size, and compare held-out with reused retrieval data under the same corpus and retrieval pipeline. This design lets us measure how external access to knowledge interacts with model capacity and pretraining saturation and quantify its likelihood-equivalent substitution for additional pretraining. 

% Prior work has extensively studied scaling laws for pretraining and, separately, the benefits of retrieval-augmented LMs, but these directions have largely been explored in isolation. We bridge this gap by studying how pretraining and retrieval interact under fixed compute and model-size constraints, with a particular focus on how to allocate data between parametric learning and external memory across different scales.

\section{Methods}
\subsection{Experimental Setup}

For our experiments, we use the OLMo-2 series \citep{olmo20242} of LMs due to its strong empirical performance, alignment with open research practices, and modern architectural design. We define our own OLMo-2 model sizes and pretraine them across various scales: 30M, 136M, 233M, 728M, 1B, and 3B parameters (hyperparameter details in Appendix~\ref{sec:appendix-pretraining}). We use 100B tokens of DCLM data as our pretraining corpus \citep{Li2025DataComp-LM:Models}. We train all models using AdamW with a $3\times10^{-4}$ peak learning rate ($lr$), $\beta_1 = 0.9, \ \beta_2 = 0.95$, and 0.1 weight decay. We adopt a warmup-stable-decay (WSD) schedule \citep{hu2024minicpm} with 10\% linear warmup (capped at 2k steps), a stable phase, and 10\% linear decay to a minimum $lr$ of $6e^{-5}$. Models are evaluated every 2k steps. % and at the end of training.

\subsection{Index Construction}
We construct retrieval indices across multiple scales (1B--20B tokens) via FAISS \citep{douze2025faiss} from both a held-out slice of DCLM and the non-held out slice that models were trained on by first computing per-chunk token counts
over the embedding store, and then selecting chunks via a seeded random permutation. For each target budget (e.g., 30M, 60M, etc.), we take the shortest prefix of that permutation whose cumulative token count meets or slightly exceeds the target, then materialize the corresponding chunk texts/metadata and build a FAISS index over the selected embeddings. Because all budgets are prefixes of the same permutation (for fixed source data, filtering config, and seed), smaller-budget indices are strict subsets of larger-budget indices (e.g., $30M \subset 60M$), enabling controlled scaling comparisons where corpus size is the primary varying factor.
For our index construction, we chose Qwen3-Embedding-8B from amongst 4 candidate choices on the basis of recall, and IVPFQ as the indexing algorithm. More index construction details are in Appendix~\ref{sec:appendix-indices}.

\begin{figure*}[t]
\centering %begin{center}
\centerline{\includegraphics[width=\textwidth]{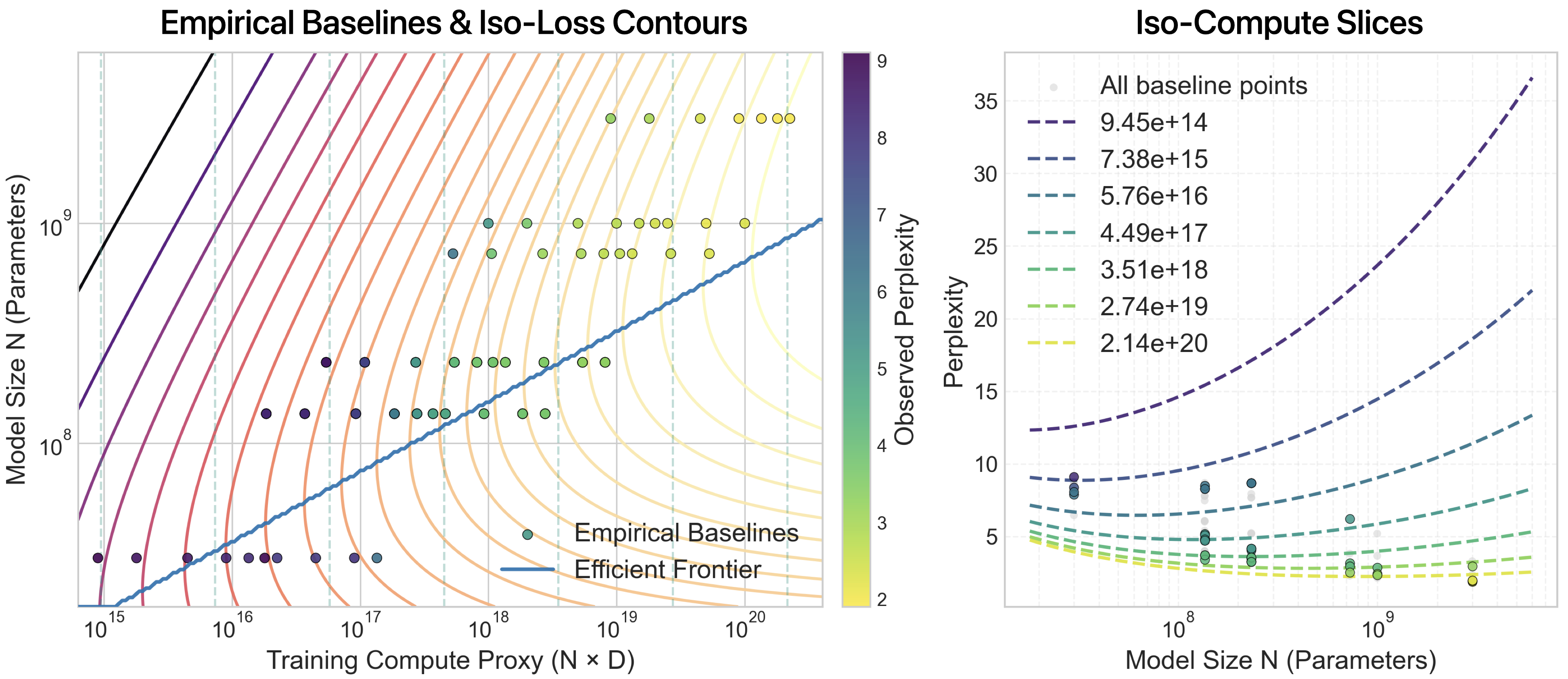}}
% \vskip 0.2cm
\caption{Parametric scaling baselines without RAG ($R=0$). \textit{Left:} Empirical measurements across model sizes and data budgets, overlaid with iso-loss contours from the power-law model. Each point corresponds to a trained model configuration, colored by observed perplexity. The blue line denotes the compute-efficient frontier and the vertical dashed lines, discrete training budgets. 
\textit{Right:} Iso-compute slices of the scaling surface, showing predicted loss as a function of model size ($N$). Empirical observations are overlaid for reference.}
\label{fig:parametric_scaling_baseline_contours}
\label{baselines}
% \vskip -0.3cm
%\end{center}
\end{figure*}

\subsection{Evaluation Protocol}
We evaluate all models using a retrieval-augmented variant of EleutherAI’s \textit{lm-evaluation-harness} \citep{eval-harness}, the \textit{RAG-Evaluation-Harness} framework \citep{Shao2024ScalingDatastore}, across multiple benchmarks spanning reasoning, scientific QA, and open-domain QA: AI2-ARC (Easy and Challenge) \citep{Clark2018ThinkChallenge}, HellaSwag \citep{Zellers2019HellaSwag:Sentence}, OpenBookQA \citep{Mihaylov2018CanAnswering}, SciQ (scientific QA) \citep{Welbl2017CrowdsourcingQuestions}, Natural Questions \citep{Kwiatkowski2019NaturalResearch}, StrategyQA \citep{Geva2021DidStrategies}, SimpleQA \citep{Wei2024MeasuringModels}, and CommonsenseQA \citep{Talmor2019CommonsenseQA:Knowledge}.

% PIQA \citep{Bisk2020PIQA:Language}

\paragraph{RAG evaluation setup.} We retrieve the top-$k$ passages ($k=5$, chosen via a small pilot sweep as a trade-off between retrieval quality and context budget) from a fixed FAISS index. %, and prepend them to the model input before the task prompt.
Retrieved passages are concatenated as context, followed by the original question and answer choices (if applicable). The retriever is frozen and shared across all evaluations to isolate the effect of retrieval scale. % and query formulation.

\paragraph{Metrics.} We evaluate retrieval at three levels. For our scaling analysis, we use the character-normalized negative log-likelihood of the gold answer continuation, conditioned on the task prompt and, for RAG, the retrieved context. We report its exponentiated value as conditional gold-answer perplexity (PPL). This continuous measure captures changes in the probability assigned to the correct answer and is better suited to fitting scaling relationships \citep{tay2021scale,krajewski2025revisiting}. Second, for multiple-choice tasks, we normalize answer-choice likelihoods into a distribution \(q\) and compute multiclass Brier loss,
\(\sum_j(q_j-\mathbf{1}[j=y])^2\). Third, decision error records whether the highest-scoring choice is incorrect (scaling fits based directly on accuracy / error-rate are included in Appendix~\ref{sec:appendix-brier}). These targets form a ladder from absolute gold-answer support, through support relative to competing answers, to the discrete decision. We use gold-answer perplexity for the primary scaling and token-allocation analyses because it varies smoothly across all benchmarks, and use Brier loss and error rate to determine whether its interactions transfer to downstream decisions.

\begin{table*}[h]
\centering
\small
\setlength{\tabcolsep}{5pt}
\begin{tabular}{llccccc}
\toprule
 & \textbf{Benchmark} & \textbf{CV ARE (\%)} & \textbf{LOMO ARE (\%)} & \boldmath{$\alpha$} & \boldmath{$\beta$} & \boldmath{$L_0$} \\
\midrule

\multirow{4}{*}{\rotatebox{90}{\textbf{\tiny Reasoning}}}
% & PIQA                   & 42.55 & 129.04 & 0.3786 & 1.3033 & 1.0920 \\
& CommonsenseQA          & 6.87  & 10.76  & 0.1558 & 0.0869 & 2.1619 \\
& HellaSwag              & 2.18  & 4.94   & 0.3923 & 0.4713 & 1.3714 \\
& StrategyQA             & 12.28 & 14.93  & 2.0000 & 0.4647 & 7.6735 \\
\midrule

\multirow{6}{*}{\rotatebox{90}{\textbf{\tiny Scientific/Open QA}}}
& SciQ   & 13.07 & 24.03 & 0.5267 & 0.2606 & 0.9522 \\
& OpenBookQA             & 4.78  & 9.41  & 0.3688 & 0.2120 & 1.6579 \\
& AI2C-ARC Easy        & 7.10  & 13.67 & 0.3566 & 0.2195 & 1.0299 \\
& AI2C-ARC Challenge   & 3.93  & 6.09  & 0.2811 & 0.2817 & 1.1133 \\
& Natural Questions      & 7.19  & 17.39 & 0.3595 & 0.3523 & 1.2593 \\
& SimpleQA               & 11.46 & 21.09 & 0.2285 & 0.9580 & 1.7196 \\
\bottomrule
\end{tabular}
\caption{
Power-law fit quality and scaling exponents for parametric baselines ($R=0$). 
We report cross-validation average relative error (CV ARE) [interpolation error under random splits] and leave-one-model-size-out ARE (LOMO ARE) [extrapolation error to unseen model size]. Lower ARE indicate better fit quality. Exponents $\alpha$ and $\beta$, when combined with $L_0$, summarize how loss scales with model size and pretraining data in the baseline regime. ARE should be interpreted relative to the inherent noise and discreteness of benchmarks, with smoother, likelihood-based tasks yielding low errors, and reasoning-heavy tasks showing higher variance and ARE.} 
\label{tab:parametric_scaling_baselines}
% \vskip -0.1in
\end{table*}

\section{Experimental Results}
\subsection{Parametric Scaling Baselines}

We begin by establishing parametric scaling baselines in the absence of retrieval (retrieval index size $R = 0$), varying model size ($N$) and pretraining data ($D$). This serves as a sanity check that our experimental setup reproduces the standard scaling-law behavior observed in prior work such as \cite{Hoffmann2022TrainingModels}. Following \cite{Hoffmann2022TrainingModels}, we model loss for parametric models as a function of model size and data using a power-law:
\begin{equation}
L(N, D) = A \left(\frac{N}{10^9}\right)^{-\alpha} + B \left(\frac{D}{10^9}\right)^{-\beta} + L_0
\label{eq:2d_baseline}
\end{equation}

where $(A,\alpha)$ capture scaling with model size, $(B,\beta)$ capture scaling with data, and $L_0$ is an irreducible loss floor. Here, $(A,\alpha)$ govern the model-size contribution, $(B,\beta)$ govern the data contribution, and $L_0$ is the asymptotic loss floor. Intuitively, larger $\alpha$ implies stronger sensitivity to model scaling, while larger $\beta$ implies stronger sensitivity to data scaling.
Across benchmarks, as seen in Figure \ref{fig:parametric_scaling_baseline_contours} and Table \ref{tab:parametric_scaling_baselines}, we observe smooth and predictable improvements as either $N$ or $D$ increases, with diminishing returns in both directions as expected from power-law scaling. The fitted model achieves low average relative error:

\vskip -0.3cm
\begin{equation*}
    \text{ARE} \,=\, \displaystyle \frac{1}{n} \sum_{i=1}^n |(\loss_i^{\text{pred}} - \loss_i^{\text{obs}}) / \loss_i^{\text{obs}}| \times 100\%
\end{equation*}

\noindent and scaling exponents broadly align with previously reported values \citep{Hoffmann2022TrainingModels}.
Overall, this validates that our setup reliably reproduces canonical scaling-law behavior.

\begin{table*}[t]
\centering
\small
\setlength{\tabcolsep}{4pt}
\begin{tabular}{llcccccccc}
\toprule
 & \textbf{Benchmark} & \textbf{CV ARE} & \textbf{LOMO}
 & \boldmath{$\alpha$} & \boldmath{$\beta$} & \boldmath{$C\!\times\!10^{3}$}
 & \boldmath{$\delta$} & \boldmath{$\zeta$} & \boldmath{$L_0$} \\
 & & (\%) & (\%) & & & & & & \\
\midrule

\multirow{3}{*}{\rotatebox{90}{\textbf{\tiny Reason.}}}
& CommonsenseQA & 5.10 & 8.77 & 0.169 & 0.087 & 27.33
  & $-0.433$ \tiny[-0.51,\,-0.33] & $-0.247$ \tiny[-0.30,\,-0.19] & 2.162 \\
& HellaSwag & 1.76 & 3.31 & 0.476 & 0.505 & 0.13
  & 1.295 \tiny[1.17,\,1.44] & 0.277 \tiny[0.26,\,0.29] & 1.484 \\
& StrategyQA & 14.33 & 18.73 & 1.178 & 0.573 & 72.10
  & $-0.109$ \tiny[-0.13,\,-0.09] & $-0.236$ \tiny[-0.25,\,-0.22] & 7.599 \\
\midrule

\multirow{6}{*}{\rotatebox{90}{\textbf{\tiny Scientific/Open QA}}}
& SciQ & 7.91 & 14.36 & 0.545 & 0.302 & 11.50
  & 0.401 \tiny[0.20,\,0.67] & 0.072 \tiny[-0.05,\,0.22] & 0.952 \\
& OpenBookQA & 3.14 & 7.13 & 0.376 & 0.239 & 4.60
  & 0.731 \tiny[0.55,\,0.97] & 0.164 \tiny[0.04,\,0.29] & 1.952 \\
& AI2-ARC Easy & 4.87 & 8.79 & 0.387 & 0.246 & 12.59
  & 0.340 \tiny[0.28,\,0.40] & 0.101 \tiny[0.07,\,0.13] & 1.030 \\
& AI2-ARC Challenge & 3.68 & 5.84 & 0.303 & 0.293 & 8.24
  & 0.469 \tiny[0.36,\,0.57] & 0.162 \tiny[0.09,\,0.23] & 1.113 \\
& Natural Questions & 5.61 & 12.76 & 0.391 & 0.392 & 13.21
  & 0.604 \tiny[0.57,\,0.63] & 0.258 \tiny[0.23,\,0.28] & 1.259 \\
& SimpleQA & 10.24 & 18.94 & 0.286 & 0.526 & 10.07
  & 0.814 \tiny[0.79,\,0.84] & 0.388 \tiny[0.36,\,0.41] & 1.720 \\
\midrule
& \textit{Mean} & \textit{6.29} & \textit{10.96} & & & & & & \\
\bottomrule
\end{tabular}
\caption{%
Per-benchmark fits of Eq.~\ref{eq:3d_scaling_law}.
\textbf{CV ARE} is the average relative error on held-out \emph{whole} $(N,D)$ retrieval curves, evaluated on retrieval points ($R>0$). \textbf{LOMO} is leave-one-model-size-out error. $\delta$ and $\zeta$ carry 95\% bootstrap intervals obtained by resampling benchmark-evaluation noise; $\delta>0$ indicates larger absolute retrieval gains for smaller models and $\zeta>0$ larger gains in less-saturated pretraining regimes. Setting $\delta=\zeta=0$ recovers an additive retrieval law in which the gain is independent of $N$ and $D$. The retrieval-rate parameter \(\eta\) reaches its upper bound of \(10^6\) on every benchmark, and profile fits continue improving or remain flat as this bound is increased. Thus, our datastore grid does not identify a finite retrieval-onset scale: the observed positive-\(R\) response is already approximately linear in \(\log R\).}

% Per-benchmark fits of the amplitude-interaction law, Eq.~\ref{eq:3d_scaling_law}. \textbf{CV ARE} is the average relative error on held-out \emph{whole} $(N,D)$ retrieval curves (5-fold grouped cross-validation; the parametric term $P(N,D)$ is refit inside each fold from training-fold $R=0$ points only, so no
% information from a held-out configuration enters the fit), evaluated on retrieval points ($R>0$). \textbf{LOMO} is leave-one-model-size-out error; it is dominated by extrapolating \emph{down} to $N=30$M (Appendix~\ref{app:lomo} gives the
% per-size breakdown). Relative to the additive law $G=C\log(1+\eta r)$, the interaction reduces CV ARE on all nine benchmarks (mean $-12.7\%$); the full nested comparison is in Table~\ref{tab:nested_retrieval_models}. $\delta$ and $\zeta$ carry 95\% bootstrap intervals obtained by resampling benchmark-evaluation noise from the reported per-configuration standard errors; $\delta>0$ indicates larger absolute retrieval gains for smaller models and $\zeta>0$ larger gains in less-saturated pretraining regimes. $C$ is reported at the convention $\eta=10^{6}$: $\eta$ is not identified by our design and must be read jointly with $C$, since only the product $C\log(1+\eta r)$ is constrained over the datastore range we cover (\S\ref{sec:identifiability}, Fig.~\ref{fig:eta_profile}). PIQA admits no usable fit under any law we tried (CV ARE $29\%$; LOMO ARE $\approx490\%$ at $N=30$M) and is deferred to Appendix~\ref{app:piqa}.
\label{tab:rag_3d_scaling_fits}
\end{table*}

\subsection{Scaling Laws for Retrieval}
\label{sec:retrieval_scaling_laws}

Our central result is that retrieval gains vary systematically with the generator’s training regime rather than contributing a fixed additive improvement. Our 3D scaling law predicts held-out retrieval curves with a mean cross-validated relative error of 6.29\% (Table~\ref{tab:rag_3d_scaling_fits}). Extrapolation across model scales is slightly less reliable: mean LOMO error rises to 10.96\%, driven primarily by extrapolation down to the 30M model (see Appendix~\ref{sec:appendix-lomo}). On seven of nine benchmarks, the fits indicate larger retrieval gains for both smaller models and models with less pretraining per parameter. These patterns remain under other seeds, and we run a stability analysis in Appendix~\ref{sec:appendix-stability}.

To capture this dependence, we extend the 2D parametric law with interactions between retrieval gain, model capacity, and pretraining exposure:
\begin{align}
L(N,D,R)
=\>
&P(N,D)-C\left(\frac{N}{10^9}\right)^{-\delta}\cdots \notag\\
&\left(\frac{D/N}{10}\right)^{-\zeta}\log\!\left(1+\eta\frac{R}{10^9}\right)
\label{eq:3d_scaling_law}
\end{align}
where $N$ is model size (parameters), $D$ is pretraining tokens, and $R$ is retrieval/index tokens. $P(N,D)$ is the parametric scaling baseline defined in Eq.~\ref{eq:2d_baseline}. While we tested several formulations for incorporating the effects of retrieval, an interaction-based law most improved cross-validation error over an additive retrieval term; we report comparisons with nested and alternative functional forms in Appendix~\ref{sec:appendix-fitting}. We evaluate predictive fit using five-fold grouped cross-validation, holding out entire $(N,D)$ retrieval curves rather than individual datastore observations. Within each fold, the parametric baseline $P(N,D)$ is refit using only the training-fold ($R=0$) configurations, ensuring that no information from a held-out model-pretraining configuration enters the fit.

Across the measured grid, a median 91\% of the largest observed retrieval gain is present at $R=N$, followed by only small improvements as the datastore grows to as much as $20N$. Therefore, we observe early near-saturation of the benefit of retrieval over the $R$ range we evaluate.

\paragraph{The interaction depends on the evaluation target.} On five of six multiple-choice benchmarks, choice-normalized Brier and decision-error fits agree in sign for both interaction exponents, whereas gold-answer likelihood and error agree on only one; their curve-bootstrap intervals are disjoint on all six (Appendix~\ref{sec:appendix-brier}). The reversal appears when competing answers are introduced: retrieval improves gold-answer likelihood more for smaller, less-pretrained models, but relative choice probability and accuracy more for larger, more-pretrained models. We therefore report them separately and restrict claims about downstream performance to the decision-level results.

% Empirically, the log-form retrieval law provides strong fits on most benchmarks — results using a power retrieval law are in Appendix \ref{sec:appendix-power-fits}. In Table~\ref{tab:rag_3d_scaling_fits}, CV ARE (Cross-Validation Average Relative Error) is low for many tasks, while LOMO (Leave-One-Model-Out) errors are generally higher, indicating that interpolation is easier than extrapolation to held-out model scales. Reasoning-heavy tasks remain less stable (especially PIQA and StrategyQA), with larger held-out errors (additional measures reported in Appendix~\ref{sec:lodo}). The fitted retrieval-rate parameter $\eta$ shows two broad regimes. For some tasks, $\eta$ is moderate ($\approx 10^{-3}$ to $\approx 2$), indicating gradual retrieval gains. For others, $\eta$ reaches the optimization ceiling (near $10$ in our current constrained fit), suggesting rapid saturation over the observed retrieval range, or limited identifiability of retrieval dynamics from available points. These results are stable across multiple training seeds (see Appendix~\ref{sec:appendix-stability}). Overall, %these fits support a consistent picture:
% retrieval improves performance with diminishing returns, and both the magnitude and saturation rate of those gains are strongly task-dependent.

\subsection{Data Reuse}

\begin{figure*}[t]
\begin{center}
\centerline{\includegraphics[width=\textwidth]{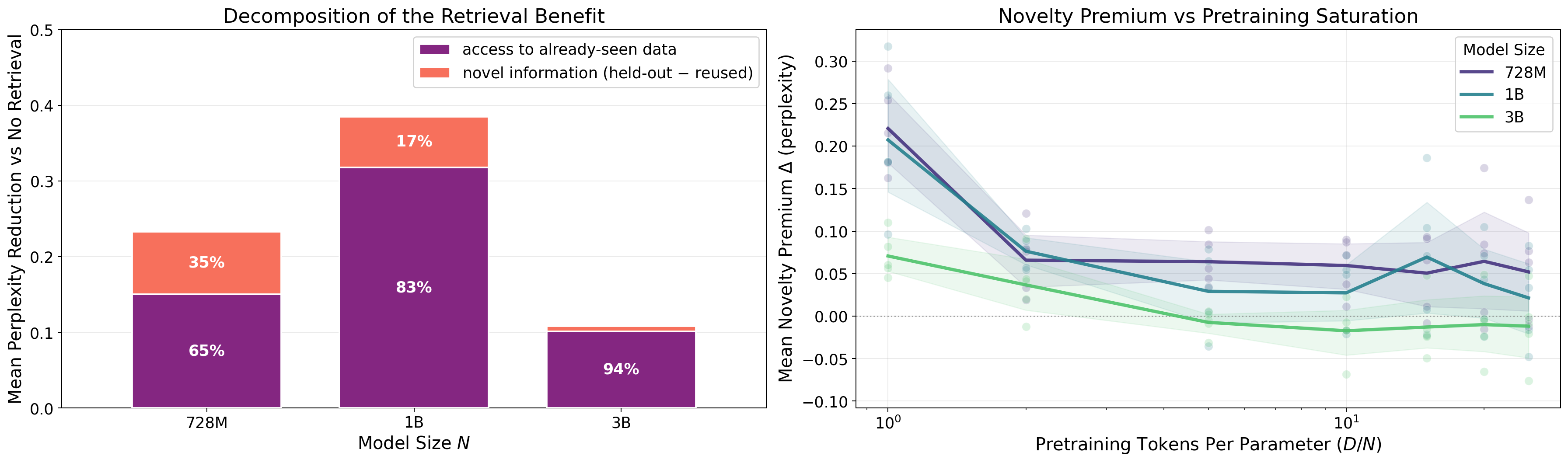}}

\caption{\textbf{Decomposing retrieval gains into reused and novel information.} \textit{Left:} Mean perplexity reduction relative to no retrieval, decomposed into the benefit from re-accessing pretraining data and the additional novelty premium from using held-out data (\textsc{Held-out} - \textsc{Reused}). Labels report each component's share of the total held-out retrieval gain. \textit{Right:} The novelty premium as a function of pretraining tokens per parameter ($D/N$). Points show benchmark-level values, lines show mean for each model size, and shaded regions show the bootstrapped 95\% confidence interval.}
\label{fig:data_reuse}
\end{center}
\vskip -0.3in
\end{figure*}

Most of retrieval's measured benefit does not require novel information: it can be achieved from making information encountered during pretraining accessible again at inference. To separate these effects, we compare, for our larger models, retrieval from held-out documents with retrieval from documents in the model's pretraining corpus. We decompose the held-out retrieval gain into an \emph{access benefit}, measured by the improvement under reuse, and a \emph{novelty premium}, measured by the additional improvement from held-out rather than reused data.

As shown in Figure~\ref{fig:data_reuse}, reuse accounts for 72-92\% of the median held-out retrieval gain for models from 728M to 3B. Thus, train--index overlap reduces the magnitude of retrieval's benefit but does not eliminate it. The novelty premium is largest in lightly pretrained regimes and generally approaches zero as $D/N$ increases. Under our protocol, retrieval therefore acts primarily as an inference-time access mechanism, while access to unseen information provides an additional benefit primarily when pretraining is limited.

\begin{figure*}[t]
\begin{center}
\centerline{\includegraphics[width=\textwidth]{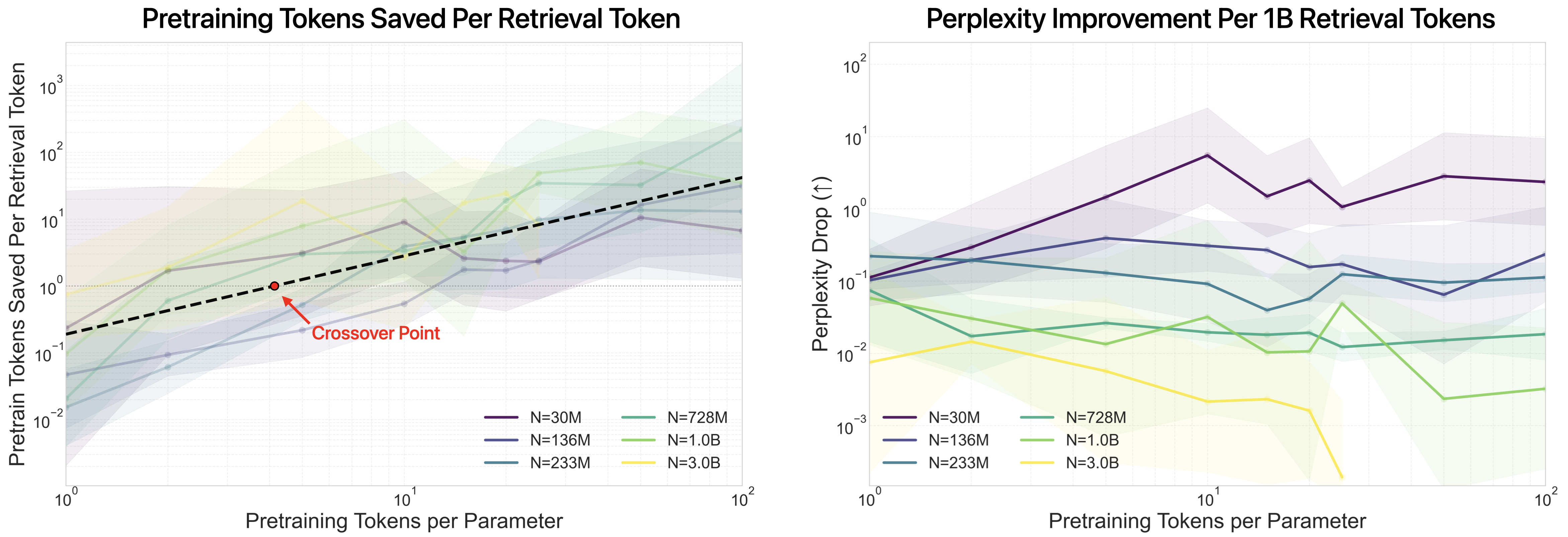}}
% \vskip 0.2cm
\caption{\textbf{Trade-off between pretraining and retrieval under a fixed data budget.} \textit{Left:} We quantify the substitutability between retrieval and pretraining via the number of pretraining tokens saved per retrieval token, computed by fitting scaling laws and determining, for each pretraining scale, the amount of retrieval required to match baseline performance without retrieval. The dotted line represents a linear line-of-best-fit across all model scales. \textit{Right:} We measure the marginal benefit of retrieval as perplexity improvement per billion retrieval tokens (higher is better) for models trained near their optimal pretraining ratio.}
\label{sigma_kappa}
\end{center}
\vskip -0.3in
\end{figure*}

\subsection{Pretraining–Retrieval Trade-off Curves}
\label{subsec:pretraining-retrieval-tradeoff}

We now investigate the trade-off between pretraining data $D$ and retrieval $R$, with the goal of understanding how retrieval can substitute for pretraining in reducing loss. Figure~\ref{sigma_kappa} summarizes this trade-off for two complementary perspectives: (i) the \emph{substitutability} of retrieval for pretraining, and (ii) the \emph{marginal benefit} of retrieval.

% trade-off across model scales. We analyze 

% \elcomment{also the takeaway seems to be that pretraining actually doesn't save that many tokens? this graph tops off at $10^3$ which is just 1k, is this correct? Also still a bit confused about the perplexity drop, up is better in the right plot right? you may want to just add ($\uparrow$)}

\paragraph{Substitutability of retrieval.} For each model and pretraining scale, we fit scaling laws and compute the amount of retrieval required to match the performance of a baseline model trained without retrieval. We express retrieval in   
units of equivalent pretraining tokens as follows. For a configuration $(N, D, R_{opt})$ with measured loss $\loss^*_{\text{RAG}}$, we project this loss onto the $N,D$ scaling curve (Eq.~\ref{eq:2d_baseline}) to find the equivalent pretraining budget, giving us the projection $D_{\text{eff}}^{\text{RAG}}$. From this, we compute the \textit{substitutability} $\sigma$ which represents the number of pretraining tokens saved per retrieval token:
\begin{align}
    D_{\text{eff}}^{\text{RAG}} &= \left( \frac{\loss^*_{\text{RAG}} - \loss_0 - A \cdot N^{-\alpha}}{B} \right)^{-1/\beta}\quad \notag\\ 
    &\Rightarrow \sigma = \frac{D_{\text{eff}}^{\text{RAG}} - D}{R_{opt}}
\end{align}
We observe a clear crossover behavior in Figure~\ref{sigma_kappa} (left). In low-data regimes, retrieval %provides limited benefit and
cannot effectively replace pretraining. However, it becomes increasingly effective beyond a median threshold of $\sim$ $D/N=5$ Pretraining Tokens Per Parameter across all model scales (estimated using the line-of-best-fit, with a 95\% confidence interval of $[2.9, \>6.4]$), with each retrieval token replacing multiple pretraining ones. In this regime, the gains grow $\sim$ log-linearly, indicating that retrieval serves as an efficient alternative to additional pretraining. Importantly, this reflects \emph{relative efficiency} rather than absolute improvement: even when retrieval substitutes efficiently for pretraining, the total achievable gain may be small if the baseline model is already near saturation.

% For a configuration $(N, D, R)$ with measured loss $\loss^*_{\text{RAG}}$:

% \begin{enumerate}
%     \item Predict baseline loss (no RAG): $\loss_{\text{baseline}} = \loss_{\text{3D}}(N, D, 0)$
%     \item Predict RAG loss: $\loss_{\text{RAG}} = \loss_{\text{3D}}(N, D, R)$
%     \item \textbf{Project both onto 2D curve}: Invert Eq.~\ref{eq:2d_baseline} to find equivalent pretraining budgets:
%     \begin{align}
%     D_{\text{eff}}^{\text{baseline}} &= \left( \frac{\loss_{\text{baseline}} - \loss_0 - A \cdot N^{-\alpha}}{B} \right)^{-1/\beta} \\
%     D_{\text{eff}}^{\text{RAG}} &= \left( \frac{\loss_{\text{RAG}} - \loss_0 - A \cdot N^{-\alpha}}{B} \right)^{-1/\beta}
%     \end{align}
%     \item Compute replacement cost:
%     \begin{equation}
%     \sigma = \frac{D_{\text{eff}}^{\text{RAG}} - D}{R}
%     \label{eq:sigma}
%     \end{equation}
% \end{enumerate}

\paragraph{Marginal benefit of retrieval} is defined as the reduction in loss per unit of retrieval data $\kappa = \displaystyle \Delta\loss / (R/10^9)$, where $\Delta\loss = \displaystyle \loss_{\text{R=0}} - \loss^*_{\text{RAG}}$ (higher is better). Figure~\ref{sigma_kappa} (right) shows this quantity for models trained near their optimal pretraining ratio. We find that smaller models (e.g., 30M) benefit most from retrieval, achieving large improvements per unit of retrieved data. As model size increases, the marginal benefit decreases, with gains diminishing substantially at larger scales and largely saturating by 3B parameters. This suggests that while retrieval may remain an efficient substitute for pretraining at larger scales (left), the absolute improvement it provides diminishes as models become increasingly saturated. We note here that efficiency refers to performance per allocated corpus token, not end-to-end compute or deployment cost, for which Appendix~\ref{sec:appendix-cost-accounting} reports a full accounting including storage and per-query inference overhead.
% At the largest datastore and pretraining configuration for each scale, estimated one-time datastore-embedding FLOPs exceed generator-pretraining FLOPs for the 30M--233M models, but fall to 0.73$\times$, 0.53$\times$, and 0.24$\times$ for the 728M, 1B, and 3B models, respectively. 

\paragraph{Summary.} There clearly exists a scale-dependent trade-off between pretraining and retrieval. As model size and pretraining increase, retrieval's marginal utility decreases, indicating a transition from retrieval-dominated to pretraining-dominated regimes. Most of the observed retrieval gain is already realized by $R=N$, and 72-92\% remains when the datastore reuses pretraining data.

% Retrieval is most valuable (a strong substitute for pretraining) in undertrained and smaller-model regimes. 

% \elcomment{hmm but I feel like figure 3 left contradicts this a bit? with more pretraining tokens RAG helps more, is this overridden by the results on figure 3 right? if the figure on the right is the main result I would demphasize the effects on the left a bit since $10^2$ pretraining toks saved is basically nothing while $10^2 $ ppl drop is good}

% \subsection{Effect of Model Scale on Optimal Split }
% does the optimal retrieval fraction shift with N?

%\kscomment{(Section) Effect of Model Scale on Optimal Split: does the optimal retrieval fraction shift with N? Not addressed in the table I suppose but this will probably have to be a short reference to the appendix.}

\subsection{Other Analyses}

Because our fixed retrieval pipeline may limit absolute gains, Appendix~\ref{sec:appendix-rag-improvements} examines alternative query formulations. Oracle-style, answer-aligned queries improve retrieval on knowledge-heavy tasks but produce little change on others, suggesting that retrieval quality shifts absolute performance without removing task-dependence.

\section{Discussion}

Our results show that retrieval and pretraining are coupled mechanisms for allocating access to corpus information. Our interaction scaling law captures this dependence: retrieval gains vary with model capacity and pretraining exposure rather than providing a fixed additive improvement. There is no metric-independent answer to which training regimes benefit most from retrieval. Retrieval improves the absolute likelihood of the gold continuation most for smaller, less-pretrained models, but improves its standing relative to competing choices, and ultimately decision accuracy, most for larger, more-pretrained models on the multiple-choice benchmarks studied. Thus, likelihood scaling characterizes how retrieved context changes support for the correct continuation, while decision-level scaling is the more relevant quantity if the objective is downstream accuracy. Gains are also strongly front-loaded: a median 91\% of the largest observed improvement occurs by $R=N$, with limited returns from further datastore growth. Retrieval is therefore a task- and regime-dependent complement to parametric learning, not a uniform substitute for pretraining.

Our findings also clarify the role of retrieval quality. Improvements from better query formulation and oracle-style retrieval indicate that some of the observed trade-off is bottlenecked by the retriever rather than the language model. However, even stronger retrieval does not eliminate the need for parametric capacity, especially on reasoning-heavy tasks where the limiting factor appears to be computation over knowledge. More broadly, this work suggests a shift in how pretraining corpora should be conceptualized. Rather than assuming that all available data should be compressed into weights, future language model design may benefit from explicitly partitioning corpora into data intended for internalization versus external access. This perspective aligns naturally with practical system design, where model capacity, training cost, memory footprint, and inference latency are all coupled.

\paragraph{Limitations.} There are several factors in our present study that could be expanded. First, the retrieval setup is intentionally simple and fixed: we use a single retriever, a fixed chunking strategy, and a fixed top-$k$ protocol. Although this isolates the effect of retrieval scale, it likely understates the gains achievable with stronger retrieval pipelines, as we briefly investigate. Second, our evaluation focuses primarily on gold-answer perplexity as the most stable metric for scaling analysis; while appropriate for fitting smooth laws, this does not fully capture all downstream behaviors of interest. Third, although we study a broad range of model sizes, our conclusions are still limited to the scales, architectures, and corpora explored here.

\paragraph{Future Work.} %These limitations point to several promising directions for future work.
Future work can test whether reranking, learned filtering, adaptive chunking, or relevance-aware retrieval changes the interactions we identify. Additionally, we fit benchmark-specific scaling laws, but another direction is to identify latent structure that explains why some tasks benefit more from retrieval than others. This maybe via characterizing benchmarks by their knowledge dependence, retrieval sensitivity, or reasoning burden, and using these properties to build more general scaling laws over model size, pretraining, and external memory. Finally, inspired by human cognition, future work could explore purposeful allocation of abstract reasoning to pretraining vs. factual knowledge to retrieval.

% \newpage
\section{Ethical Considerations}

Our experiments require non-trivial computation for model pretraining, datastore embedding, and evaluation, with associated energy use and environmental impact. We limit experiments to models no larger than 3B parameters, reuse a fixed embedding model and precomputed embeddings across conditions, and construct nested datastores to avoid redundant processing. Appendix~\ref{sec:appendix-cost-accounting} provides an estimated accounting of these computational costs. Additionally, our models and retrieval stores are derived from DCLM web data, which may contain social biases, offensive material, personal information, or copyrighted text. Retrieval can make such source content more directly accessible at inference time. Our experiments study aggregate scaling behavior rather than deploying a user-facing system, but these risks should be considered before applying the resulting methods in practice.

% \section*{Acknowledgments}

% ACL Anthology + Custom bibliography entries only
% anthology-1, anthology-2
\bibliography{custom,babylm_refs}

\newpage
\appendix

\section{Appendix}
\label{sec:appendix}

\subsection{Pretraining setup}
\label{sec:appendix-pretraining}

All pretraining runs use NVIDIA H100 GPUs with 8 devices per job. The total expenditure across training, retrieval embedding and index creation, and evaluation was approximately $15,000$ GPU hours. We train with FSDP data-parallelism, in mixed precision, and without model parallelism. We use varied per-device micro-batch sizes depending on the model scale, and gradient accumulation to achieve an effective global batch size of 256 across runs. Training uses context length (block size) 4096 across all models. Additional hyperparameter details for each model scale is provided in Table~\ref{tab:scaling_hparams}.

\subsection{Index Construction}
\label{sec:appendix-indices}
\subsubsection{Retrieval Corpus}
\label{subsubsec:appendix-indices-part1}
We construct a retrieval corpus from a held-out split of the DCLM dataset, chunked into overlapping token windows and embedded using a pretrained embedding model. All embeddings are $L2$-normalized and indexed using FAISS with product quantization. We vary the retrieval corpus size across multiple scales and construct separate indices for each setting. Each split corresponds to an increasing retrieval corpus size derived from disjoint subsets of the DCLM corpus. Index construction-relevant details are in Tables \ref{tab:app_index_1}, \ref{tab:app_index_2}, and \ref{tab:app_index_3}. %\sfcomment{emmy, karan, and michael: is this section done?}
\subsubsection{Embedding Choice, Chunking \& Tokenization}
\label{subsubsec:appendix-indices-part2}
We explored several embedding models including BAAI/bge-base-en-v1.5, google/embeddinggemma-300m, and Qwen3-Embedding-8B. We chose Qwen3-Embedding-8B as it showed strongest semantic recall and was consistently at the top of public RAG benchmarks. We used IVFPQ for our indexing algorithm, a chunk length of 900 tokens, and a stride length of 256 tokens. We built our chunks using TikToken cl100k-base \citep{tiktoken}, then decoded back to text. Tables~\ref{tab:app_index_1}, \ref{tab:app_index_2}, and \ref{tab:app_index_3} describe our specific embedding, chunking, and FAISS configurations.

\begin{table}[h!]
\centering
\small
\begin{tabular}{ll}
\toprule
\textbf{Component} & \textbf{Configuration} \\
\midrule
Embedding model & Qwen3-Embedding-8B \\
Embedding dimension & 4096 \\
Pooling & Last non-padded token \\
Normalization & L2-normalized \\
Tokenizer & cl100k\_base \\
Chunk length & 900 tokens \\
Stride & 256 tokens ($\sim$28\% overlap) \\
\bottomrule
\end{tabular}
\caption{Embedding and chunking configuration for retrieval corpus construction.}\label{tab:app_index_1}
\end{table}

\begin{table}[h!]
\centering
\small
\begin{tabular}{ll}
\toprule
\textbf{Component} & \textbf{Configuration} \\
\midrule
Index type & IVFPQ (inner product) \\
Sub-quantizers & 128 \\
Bits per code & 8 \\
Search parameter & nprobe = 64 \\
Training set size & $\max(\text{nlist} \times 50, 10^5)$ \\
Batch size (add) & 500K vectors \\
\bottomrule
\end{tabular}
\caption{FAISS index configuration.}\label{tab:app_index_2}
\end{table}

\begin{table}[h!]
\centering
\small
\begin{tabular}{ccc}
\toprule
\textbf{Corpus Split} & \textbf{nlist} & \textbf{Shards} \\
\midrule
5\% & 32,768 & 32 \\
10\% & 32,768 & 64 \\
20\% & 65,536 & 128 \\
30\% & 65,536 & 128 \\
\bottomrule
\end{tabular}
\caption{Per-split FAISS index configuration.}
\label{tab:app_index_3}
\end{table}

\begin{table*}[htb!]
\centering
\label{tab:scaling_hparams}
\resizebox{\textwidth}{!}{\begin{tabular}{c c c c c c c}
\hline
Model Size & Layers & Hidden Dim & Number of Heads & Query Groups & Intermediate Size & Block Size \\
\hline
30M   & 8   & 256  & 4 & 4 & 512 & 4096 \\
136M  & 8   & 512  & 8 & 8 & 2048 & 4096 \\
233M  & 16  & 640  & 10 & 10 & 2560 & 4096 \\
728M  & 18  & 1280 & 10 & 10 & 5120 & 4096 \\
1B    & 24  & 1408 & 11 & 11 & 5632 & 4096 \\
3B    & 26  & 2560 & 20 & 20 & 10240 & 4096 \\
\hline
\end{tabular}}
\caption{Pretraining hyperparameters across model sizes}\label{tab:scaling_hparams}
\end{table*}

\subsection{Cost Accounting}
\label{sec:appendix-cost-accounting}

We estimate generator pretraining as \(F_{\mathrm{train}}\approx6ND\) and one-time datastore embedding as \(F_{\mathrm{embed}}\approx2N_eR\), where \(N_e=8\)B for Qwen3-Embedding-8B. Table~\ref{tab:cost-accounting} reports these costs at the largest pretraining and retrieval scales evaluated for each generator. Because the datastores are nested, embedding cost is charged once for the largest store rather than separately for every datastore size.

\begin{table*}[t]
\centering
\small
\setlength{\tabcolsep}{6pt}
\begin{tabular}{lrrrrr}
\toprule
\textbf{Generator}
& \boldmath{\(D_{\max}\)}
& \boldmath{\(R_{\max}\)}
& \boldmath{\(F_{\mathrm{train}}\)}
& \boldmath{\(F_{\mathrm{embed}}\)}
& \textbf{Embed / Train} \\
& \textbf{(B tokens)}
& \textbf{(B tokens)}
& \textbf{(FLOPs)}
& \textbf{(FLOPs)}
& \\
\midrule
30M  & 4.50   & 0.60  & \(8.10{\times}10^{17}\) & \(9.60{\times}10^{18}\) & 11.85 \\
136M & 20.40  & 2.72  & \(1.66{\times}10^{19}\) & \(4.35{\times}10^{19}\) & 2.61 \\
233M & 34.95  & 4.66  & \(4.89{\times}10^{19}\) & \(7.46{\times}10^{19}\) & 1.53 \\
728M & 72.80  & 14.56 & \(3.18{\times}10^{20}\) & \(2.33{\times}10^{20}\) & 0.73 \\
1B   & 100.00 & 20.00 & \(6.00{\times}10^{20}\) & \(3.20{\times}10^{20}\) & 0.53 \\
3B   & 75.00  & 20.00 & \(1.35{\times}10^{21}\) & \(3.20{\times}10^{20}\) & 0.24 \\
\bottomrule
\end{tabular}
\caption{Approximate generator-pretraining and one-time datastore-embedding costs at the largest evaluated scales. Estimates use \(6ND\) for training and \(2N_eR\) for embedding with \(N_e=8\)B.}
\label{tab:cost-accounting}
\end{table*}

Embedding is therefore more expensive than generator pretraining for the 30M--233M models, but falls to \(0.73\times\), \(0.53\times\), and \(0.24\times\) of pretraining cost for the 728M, 1B, and 3B models. The fixed embedder does not confound comparisons among datastore sizes for a given generator, but its absolute cost is non-negligible, particularly for small generators.

Retrieval also introduces storage and per-query overhead. Under our IVFPQ configuration, the compressed vector codes for a 20B-token store require approximately 3.0GB, plus approximately 1.1GB for coarse centroids; storing the associated tokenized text can require up to 80GB. For a representative 128-token query, query embedding costs approximately \(2.05{\times}10^{12}\) FLOPs. Retrieving five 900-token chunks can fill the generator's 4,096-token context window, adding approximately \(1.02{\times}10^{13}\) and \(2.83{\times}10^{13}\) prefill FLOPs for the 1B and 3B generators, respectively. At one million queries, query embedding and additional prefill amount to approximately \(2.0\%\) and \(2.2\%\) of their largest pretraining runs.

\begin{table*}[t]
\centering
\footnotesize
\setlength{\tabcolsep}{4pt}
\begin{tabular}{llcccccccc}
\toprule
 & \textbf{Benchmark} & \textbf{CV ARE} & \textbf{LOMO}
 & \boldmath{$\alpha$} & \boldmath{$\beta$} & \boldmath{$C\!\times\!10^{3}$}
 & \boldmath{$\delta$} & \boldmath{$\zeta$} & \boldmath{$L_0$} \\
 & & (\%) & (\%) & & & & & & \\
\midrule
\multirow{2}{*}{\rotatebox{90}{\textbf{\tiny Reas.}}}
& CommonsenseQA & 4.10 & 10.53 & 0.157 & 0.098 & 1.06
  & $-1.462$ \tiny[$-$1.65,\,$-$1.30] & $-0.699$ \tiny[$-$0.80,\,$-$0.60] & 0.255 \\
& HellaSwag & 4.16 & 7.81 & 0.100 & 0.069 & 0.40
  & $-2.000$\rlap{$^{\ddagger}$} \tiny[$-$2.00,\,$-$2.00] & $-0.924$ \tiny[$-$0.97,\,$-$0.88] & 0.262 \\
\midrule
\multirow{4}{*}{\rotatebox{90}{\textbf{\tiny Scientific QA}}}
& SciQ & 17.30 & 51.96 & 0.618 & 0.359 & 0.34
  & $-1.216$ \tiny[$-$1.51,\,$-$0.99] & $-0.751$ \tiny[$-$0.97,\,$-$0.62] & 0.046 \\
& OpenBookQA & 2.23 & 4.95 & 0.068 & 0.047 & 1.69
  & $-1.112$ \tiny[$-$1.49,\,$-$0.85] & $-0.595$ \tiny[$-$0.80,\,$-$0.42] & 0.364 \\
& AI2-ARC Easy & 7.27 & 20.32 & 0.259 & 0.171 & 2.32
  & $-0.894$ \tiny[$-$0.98,\,$-$0.82] & $-0.474$ \tiny[$-$0.52,\,$-$0.43] & 0.148 \\
& AI2-ARC Challenge & 2.74 & 5.85 & 0.056 & 0.036 & 0.99
  & $-1.524$ \tiny[$-$1.75,\,$-$1.34] & $-0.817$ \tiny[$-$0.94,\,$-$0.71] & 0.328 \\
\midrule
& \textit{Mean} & \textit{6.30} & \textit{16.90} & & & & & & \\
\bottomrule
\end{tabular}
\caption{
Eq.~\ref{eq:3d_scaling_law} refit to \textbf{downstream error rate}
($L = 1 - \mathrm{accuracy}$) instead of gold-answer perplexity, on the six multiple-choice benchmarks for which accuracy is defined and produces stable fits.}
\label{tab:rag_3d_scaling_fits_accuracy}
\end{table*}

\begin{figure*}[t]
\begin{center}
\centerline{\includegraphics[width=\textwidth]{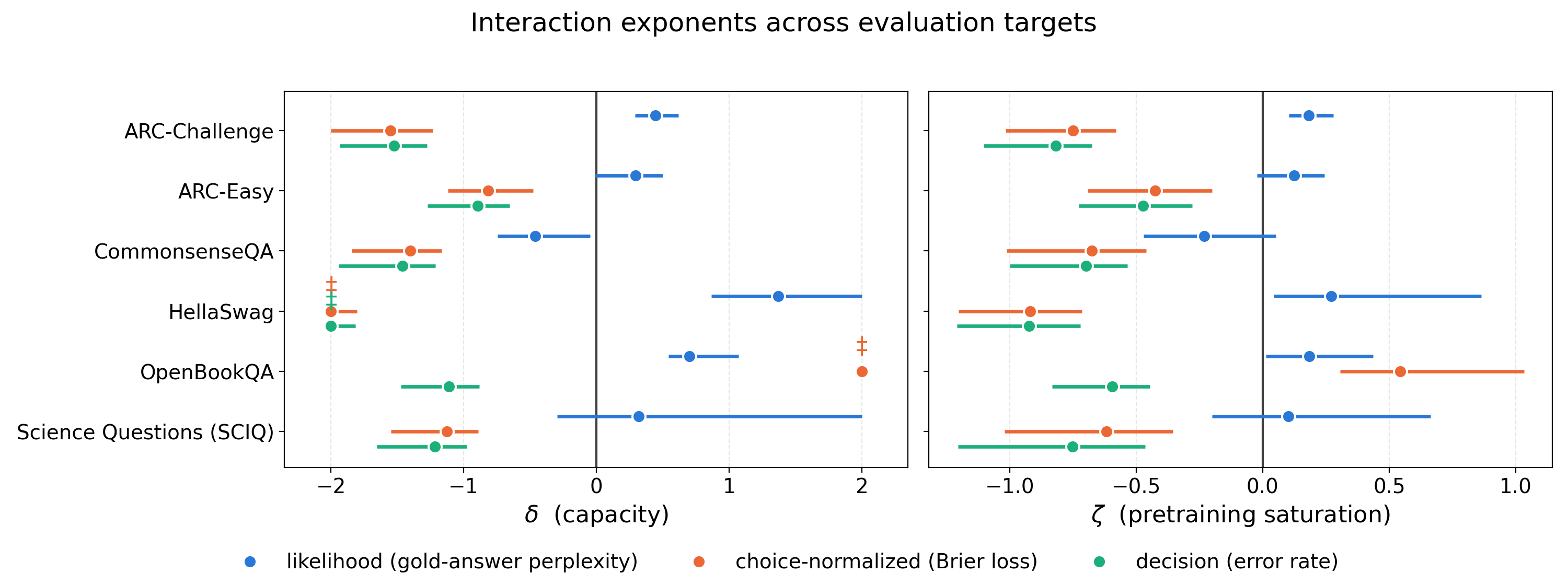}}
\caption{\textbf{Interaction exponents across evaluation targets.} Fitted capacity exponent $\delta$ (left) and pretraining-exposure exponent $\zeta$ (right) under gold-answer perplexity, choice-normalized Brier loss, and decision error. Positive values indicate larger retrieval gains for smaller models ($\delta$) or models with fewer pretraining tokens per parameter ($\zeta$). Bars show 95\% intervals obtained by resampling complete $(N,D)$ retrieval curves. Estimates at $\pm2$ lie on an optimization bound.}
\label{fig:metrics_forest}
\vskip -0.6cm
\end{center}
\end{figure*}

\subsection{Choice-Normalized and Accuracy Fits}
\label{sec:appendix-brier}

Our primary scaling laws use gold-answer perplexity, which measures the model's absolute support for the correct continuation without reference to competing choices. To determine whether the resulting interactions transfer to downstream decisions, we refit Eq.~\ref{eq:3d_scaling_law} using two additional targets on the six multiple-choice benchmarks: choice-normalized Brier loss and decision error. Brier loss is computed after normalizing likelihoods across the answer choices, while decision error applies the final argmax. These targets form a ladder from absolute gold-answer support, through competition among choices, to the discrete prediction.

% We refit Eq.~\ref{eq:3d_scaling_law} using downstream error rate ($L=1-\mathrm{accuracy}$) on the six benchmarks with stable accuracy measurement, with results shown in Table~\ref{tab:rag_3d_scaling_fits_accuracy}. The interaction law achieves a mean CV ARE of (6.30\%), nearly identical to that of the law fit on perplexity. The direction of the interaction, however, differs from the likelihood result. Both $\delta$ and $\zeta$ are negative on all six benchmarks, with nominal bootstrap intervals excluding zero. This indicates larger retrieval gains for larger and more heavily pretrained models -- the opposite pattern from the likelihood fits on five of the six shared benchmarks. We omit benchmarks without stable accuracy fits. Natural Questions and SimpleQA exhibit near-constant exact-match error, while StrategyQA drives interaction parameters to their optimization bounds.

The decision-error law achieves (6.30\%) mean CV ARE, comparable to the perplexity law's (6.29\%) (Table~\ref{tab:rag_3d_scaling_fits_accuracy}). However, both interaction exponents are negative on all six accuracy benchmarks, indicating larger decision-level gains for larger and more heavily pretrained models. This reverses the likelihood pattern on five of the six shared benchmarks. We omit benchmarks without stable decision-level fits. Natural Questions and SimpleQA have nearly constant exact-match error, while StrategyQA drives interaction parameters to their bounds. 

Figure~\ref{fig:metrics_forest} localizes where this reversal appears. For both exponents, Brier and decision-error fits agree in sign on five of six benchmarks, whereas perplexity and decision error agree on only one. The perplexity and decision-error intervals are disjoint on all six benchmarks for both exponents, indicating that the difference is not explained by uncertainty under the curve-resampling procedure. The interaction therefore changes when competing answers enter the evaluation, before the final argmax: retrieval improves absolute gold-answer likelihood most for smaller, less-pretrained models, but improves the gold answer relative to its competitors most for larger, more-pretrained models. The metrics therefore characterize distinct scaling targets. Gold-answer perplexity measures how retrieved context changes support for the correct continuation; decision errors measure whether that support improves relative to alternative answers.

\begin{table*}[t]
\centering
\small
\setlength{\tabcolsep}{5pt}
\begin{tabular}{lcccccccc}
\toprule
\textbf{Benchmark} & \textbf{30M} & \textbf{136M} & \textbf{233M}
 & \textbf{728M} & \textbf{1B} & \textbf{3B}
 & \textbf{All folds} & \textbf{Excl.\ 30M} \\
\midrule
CommonsenseQA      & 19.6 &  4.4 &  3.7 &  4.1 &  5.2 & 15.5 &  8.8 &  6.6 \\
HellaSwag          & 10.2 &  1.5 &  2.6 &  1.7 &  1.7 &  2.7 &  3.4 &  2.1 \\
StrategyQA         & 93.5 & 16.3 & 14.4 & 17.3 &  7.2 &  7.6 & 26.1 & 12.6 \\
\midrule
SciQ               & 28.8 & 10.3 &  8.9 &  7.1 &  7.4 &  2.7 & 10.9 &  7.3 \\
OpenBookQA         & 18.4 &  2.7 &  3.9 &  3.8 &  4.1 &  5.6 &  6.4 &  4.0 \\
AI2-ARC Easy       & 18.4 &  8.1 &  4.7 &  4.1 &  6.7 &  3.1 &  7.5 &  5.4 \\
AI2-ARC Challenge  & 11.7 &  4.7 &  2.7 &  3.1 &  4.5 &  3.0 &  4.9 &  3.6 \\
Natural Questions  & 54.1 &  7.2 &  3.4 &  4.5 &  6.8 &  4.5 & 13.4 &  5.3 \\
SimpleQA           & 46.3 & 16.4 &  7.4 &  5.2 & 10.2 &  3.4 & 14.8 &  8.5 \\
\midrule
\textit{Mean}      & \textit{33.4} & \textit{8.0} & \textit{5.8} & \textit{5.7}
                   & \textit{6.0} & \textit{5.3} & \textit{10.7} & \textit{6.1} \\
\bottomrule
\end{tabular}
\caption{%
LOMO average relative error (\%) on held-out retrieval points ($R>0$), by which model size was withheld.%
}
\label{tab:lomo_breakdown}
\end{table*}

\subsection{Leave-One-Model-Out Errors}
\label{sec:appendix-lomo}

Leave-one-model-out (LOMO) validation removes one model size entirely, refits the law (both the parametric term and the retrieval term) on the remaining five, and predicts the held-out size's retrieval curves. It therefore measures extrapolation across scales, unlike the grouped cross-validation of Table~\ref{tab:rag_3d_scaling_fits}, which holds out whole $(N,D)$ retrieval curves but never an entire model size.

As Table~\ref{tab:lomo_breakdown} shows, LOMO error is concentrated in the 30M fold. Withholding 30M produces a 33.4\% mean error, compared with 5.3-8.0\% for every other held-out size. Excluding this fold reduces the mean to 6.1\%, close to the grouped-CV error. This fold uniquely requires downward extrapolation from models spanning 136M--3B. Under the power-law baseline, uncertainty in the capacity exponent is amplified toward smaller (N), while the fit must also predict the large retrieval responses observed at 30M. The resulting errors are concentrated on StrategyQA (93.5\%), Natural Questions (54.1\%), and SimpleQA (46.3\%), which exhibit the noisiest behavior at the smallest scale.

By contrast, withholding the upper boundary at 3B produces the lowest mean error, 5.3\%. The elevated aggregate LOMO error therefore reflects difficulty extrapolating below the observed range rather than uniformly weak transfer across model sizes. Within the scales studied, interpolation and upward prediction to 3B are comparable in accuracy to predicting held-out retrieval curves.

\subsection{Additional Forms}
\label{sec:appendix-fitting}

We compare alternative retrieval laws to determine whether retrieval is additive, which aspects of the training regime modulate its benefit, and whether the observed datastore response supports a more flexible functional form. All models share the same parametric baseline $P(N,D)$, satisfy $G(R=0)=0$, and are evaluated using the grouped cross-validation protocol described in Section~\ref{sec:retrieval_scaling_laws}. 
% We define $n=N/10^9$, $s=(D/N)/10$, and $r=R/10^9$.

Table~\ref{tab:nested_retrieval_models} first establishes that a retrieval term is necessary: the no-retrieval model has substantially higher predictive error than the additive law $M_0$. Allowing retrieval magnitude to depend on model capacity alone reduces mean CV ARE from 7.04\% to 6.67\%, whereas conditioning only on pretraining saturation yields a smaller reduction to 6.98\%. Conditioning on both produces the strongest nested model, $M_{NS}$, which reaches 6.29\% CV ARE and improves over $M_0$ on all benchmarks. This comparison supports model capacity as the dominant interaction, with $D/N$ providing a weaker complementary effect.

More flexible laws do not yield a clear improvement in out-of-scale prediction. Allowing $N$ and $D/N$ to modulate the retrieval rate lowers interpolation error, as do Hill laws with a varying transition point, but both increase LOMO error relative to $M_{NS}$. Their rate or transition parameters also lie on optimization bounds in nearly every benchmark. The bounded Hill interaction $H_{NS}$ performs nearly identically to the logarithmic $M_{NS}$, showing that the magnitude-interaction result is not specific to the logarithmic response. Replacing absolute datastore size with $R/N$ also leaves both CV and LOMO error essentially unchanged. We select $M_{NS}$ as the primary law because it captures the supported interaction with fewer assumptions than the rate-varying alternatives.

\begin{table*}[t]
\centering
\small
\setlength{\tabcolsep}{5pt}
\begin{tabular}{llccccccc}
\toprule
\textbf{Model} & \textbf{Retrieval gain} \boldmath{$G$} & \textbf{\#par.}
 & \textbf{CV ARE} & \boldmath{$\Delta$} & \textbf{wins}
 & \textbf{LOMO} & \boldmath{$\Delta$} \\
 & & & (\%) & (\%) & & (\%) & (\%) \\
\midrule
$M_{\emptyset}$ & $0$ (no retrieval term) & 0 & 9.24 & $+32.5$ & 0/9 & 14.91 & $+18.1$\\
\midrule
$M_0$ & $C\log(1+\eta r)$ & 2 & 7.04 & --- & --- & 12.85 & --- \\
$M_N$ & $C n^{-\delta}\log(1+\eta r)$ & 3 & 6.67 & $-7.2$ & 8/9 & 11.36 & $-12.7$ \\
$M_S$ & $C s^{-\zeta}\log(1+\eta r)$ & 3 & 6.98 & $-1.4$ & 7/9 & 12.82 & $-0.5$\\
\textbf{$M_{NS}$} & $C n^{-\delta}s^{-\zeta}\log(1+\eta r)$ & 4
  & \textbf{6.29} & \boldmath{$-12.7$} & \textbf{9/9} & \textbf{10.96} & \boldmath{$-16.3$} \\
$M_{NS,\eta}$ & $C n^{-\delta}s^{-\zeta}\log(1+\eta n^{-u}s^{-v} r)$ & 6
  & 6.06 & $-16.3$ & 9/9 & 11.42 & $-12.0$ \\
\midrule
% $H_0$ & $C\,r/(r+K)$ & 2 & 7.01 & $-0.5$ & 7/9 & 12.83 & $-0.2$ & 8/9 \\
$H_{NS}$ & $C n^{-\delta}s^{-\zeta}\,r/(r+K)$ & 4 & 6.30 & $-12.7$ & 9/9 & 11.23 & $-13.8$ \\
$H_{NS,K}$ & $C n^{-\delta}s^{-\zeta}\,r/(r+K_0 n^{\kappa}s^{\lambda})$ & 6
  & 5.94 & $-17.8$ & 9/9 & 11.50 & $-10.0$\\
\midrule
$M_{NS}$ ($R/N$) & as $M_{NS}$, with $r\!\to\!R/N$ & 4 & 6.28 & $-12.9$ & 9/9 & 10.94 & $-16.1$\\
\bottomrule
\end{tabular}
\caption{\textbf{Comparison of retrieval-gain specifications.} Results are averaged across the benchmarks. All specifications share the same five-parameter baseline $P(N,D)$. $\Delta$ is the relative change from the additive law $M_0$ and \textbf{wins} counts benchmarks with lower grouped-CV error than $M_0$.}
\label{tab:nested_retrieval_models}
\end{table*}

\begin{table*}[h]
\centering
\small
\begin{tabular}{lcccc|cccc}
\toprule
\textbf{Benchmark}
& \textbf{Mean $\sigma$}
& $\sigma@1\times$
& $\sigma@10\times$
& $\sigma@100\times$
& \textbf{Mean $\kappa$}
& $\kappa@1\times$
& $\kappa@10\times$
& $\kappa@100\times$ \\
\midrule
AI2C-ARC Challenge
& 4.73 & 0.06 & 9.14 & 11.20
& 0.4778 & 0.1582 & 2.4505 & 0.4216 \\

AI2C-ARC Easy
& 22.54 & 0.38 & 27.83 & 64.40
& 0.3701 & 0.1048 & 0.4348 & 0.5736 \\

CommonsenseQA
& 677.51 & 677.93 & 1216.95 & 1951.00
& 0.5738 & 0.2715 & 1.4521 & 1.0438 \\

NQ-Open
& 5.28 & 0.23 & 2.72 & 13.63
& 0.4421 & 0.4118 & 0.5427 & 0.3192 \\

OpenBookQA
& 6.65 & 0.32 & 6.05 & -10.49
& 2.2450 & 0.2014 & 3.8579 & 0.4100 \\

SciQ
& 24.21 & 0.20 & 11.69 & 78.62
& 0.9830 & 0.3722 & 0.2576 & 0.4918 \\

StrategyQA
& 95.59 & 7.89 & 0.95 & 105.40
& 3.5006 & 0.0420 & 0.3923 & 2.1839 \\
\bottomrule
\end{tabular}
\caption{Replacement cost $\sigma$ and marginal benefit $\kappa$
across benchmarks and training regimes.}
\label{tab:app-sigma-kappa}
\end{table*}

\subsection{Some Further Notes On Retrieval Efficiency Metrics ($\sigma$ and $\kappa$)}\label{sec:appendix-retrieval-efficiency-metrics} 
%\sfcomment{refer to this in the body}

To complement the scaling-law analysis in \S\ref{subsec:pretraining-retrieval-tradeoff}, we introduce two derived metrics that quantify the efficiency of retrieval relative to pretraining: \textbf{replacement cost} ($\sigma$) and \textbf{marginal benefit} ($\kappa$). These metrics provide an interpretable view of the trade-off between parametric and non-parametric knowledge. We explain them further here.

\paragraph{Multiplicative nature of $\sigma$.}
Replacement cost $\sigma$ is a ratio that measures how many pretraining tokens are replaced per retrieval token. As such, it is inherently multiplicative: a change from $\sigma = 1$ to $\sigma = 10$ represents a tenfold increase in efficiency.

Consistent with this interpretation, $\sigma$ spans multiple orders of magnitude across tasks (e.g., from $<1$ to $>10^3$), and follows approximately log-linear trends with respect to pretraining scale. Therefore, we aggregate $\sigma$ using the geometric mean:
\begin{equation}
\sigma_{\text{GM}} = \exp\left( \frac{1}{n} \sum_{i=1}^n \ln \sigma_i \right)
\end{equation}

This preserves multiplicative structure, reduces sensitivity to extreme values, and aligns with the power-law scaling behavior underlying our analysis.

\paragraph{Additive nature of $\kappa$.}
In contrast, the marginal benefit $\kappa$ measures an absolute reduction in loss per unit of retrieval data. This is an additive quantity: improvements combine linearly and may be positive, zero, or negative depending on the task.

Because $\kappa$ can take non-positive values and does not exhibit multiplicative structure, geometric aggregation is not appropriate. Instead, we summarize $\kappa$ using the median:
\begin{equation}
\kappa_{\text{med}} = \text{median}(\{\kappa_1, \ldots, \kappa_n\})
\end{equation}
which provides a robust estimate of the typical improvement across tasks.

\paragraph{Pretraining Regimes.}
We report results across three regimes defined by the token-to-parameter ratio:
\begin{itemize}[topsep=0pt,leftmargin=8mm,itemsep=0pt,partopsep=0pt,parsep=2pt]
    \item $1\times$: $D/N \approx 1$ (undertrained)
    \item $10\times$: $D/N \approx 10$ (near-optimal)
    \item $100\times$: $D/N \approx 100$ (overtrained)
\end{itemize}

\begin{figure*}[t]
\begin{center}
\centerline{\includegraphics[width=\textwidth]{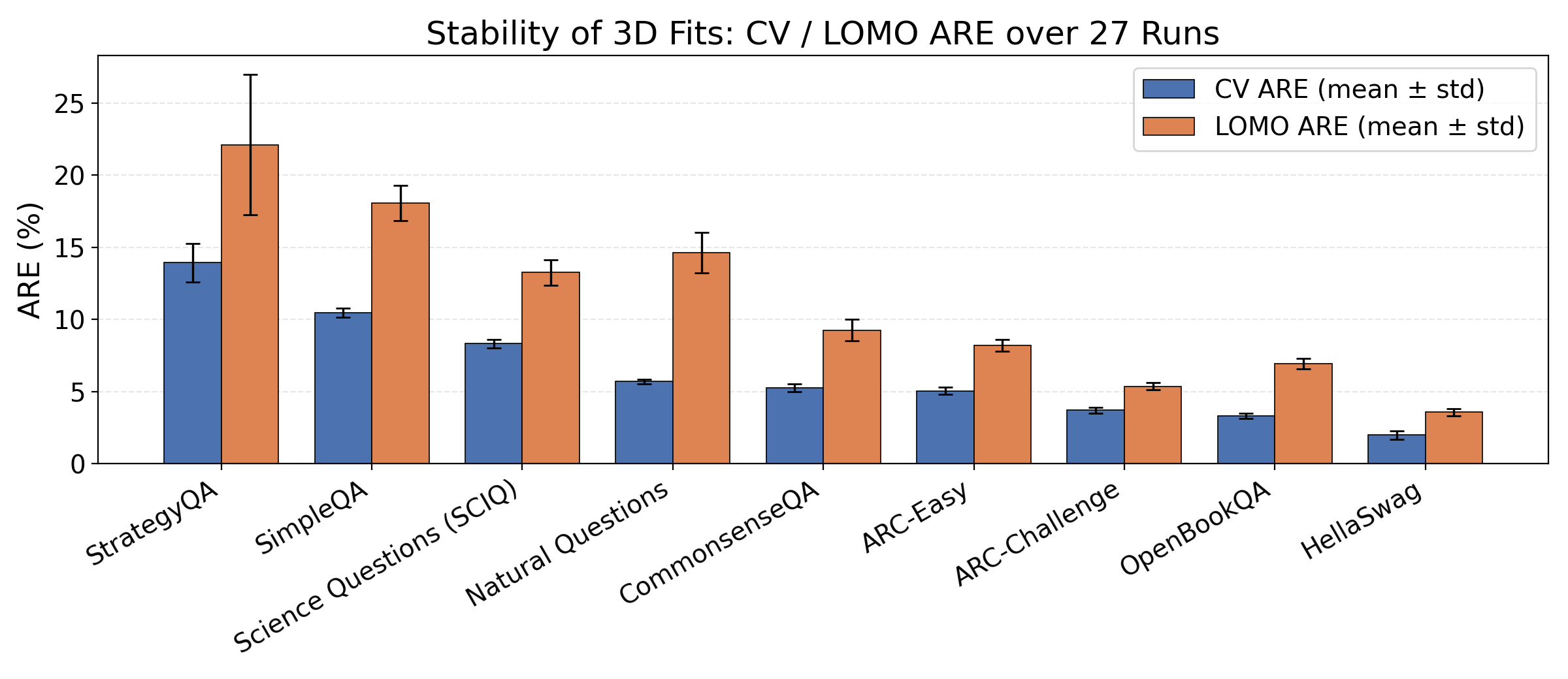}}
\vskip 0.2cm
\caption{
\textbf{Stability of scaling-law fits across random seeds.}
We report mean $\pm$ standard deviation of cross-validation ARE (CV ARE) and leave-one-model-out ARE (LOMO ARE) across 27 separate fits (every possible combination of 3 model families $\times$ 3 seeds each).
}
\label{fig:seed_stability}
\end{center}
\end{figure*}

\subsubsection{Benchmark-Level Results}

Table \ref{tab:app-sigma-kappa} shows $\sigma$ and $\kappa$ at various training scales for each benchmark. 

\paragraph{Observations:}
\begin{itemize}[topsep=0pt,leftmargin=8mm,itemsep=0pt,partopsep=0pt,parsep=2pt]
    \item \textbf{Retrieval substitutability increases with saturation:} $\sigma$ grows with $D$, indicating that retrieval becomes more effective once pretraining enters diminishing returns.
    \item \textbf{Diminishing marginal returns:} $\kappa$ decreases with model size, consistent with larger models internalizing more knowledge parametrically.
    \item \textbf{Strong task dependence:} knowledge-intensive tasks (e.g., CommonsenseQA) exhibit very high $\sigma$, while reasoning-heavy tasks (e.g., HellaSwag) show weak or negative gains.
\end{itemize}

These results provide a quantitative interpretation of the trade-off curves in Figure \ref{sigma_kappa}, reinforcing the view that retrieval acts as a scale-dependent substitute for pretraining.

\subsection{Calibration Plots}
\label{subsec:appendix-calibration-plots}
\begin{figure*}[t]
    \centering
    % Row 1
    \begin{subfigure}[b]{0.49\textwidth}
        \centering
        \includegraphics[width=\textwidth]{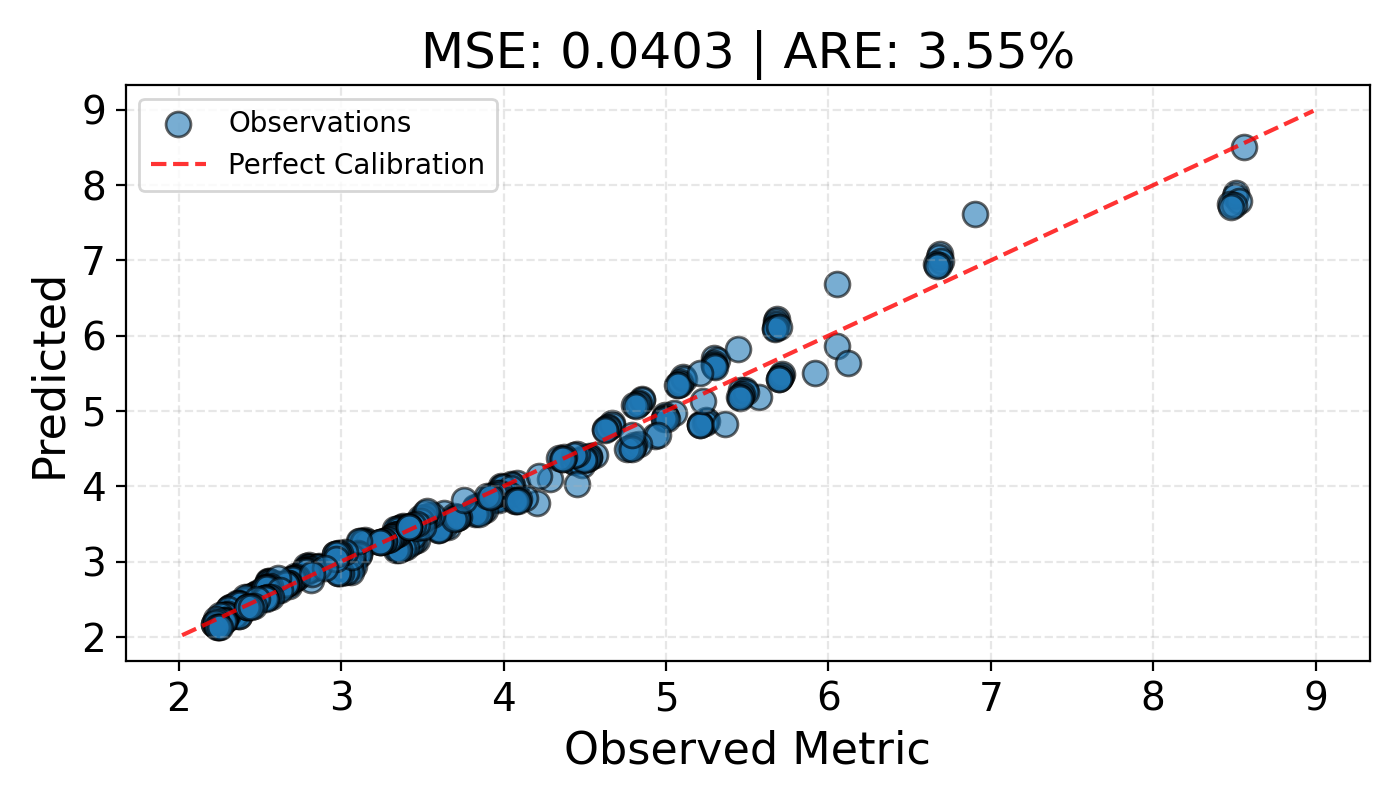}
        \caption{ARC Challenge}
        \label{fig:calib_arc_challenge}
    \end{subfigure}
    \hfill
    \begin{subfigure}[b]{0.49\textwidth}
        \centering
        \includegraphics[width=\textwidth]{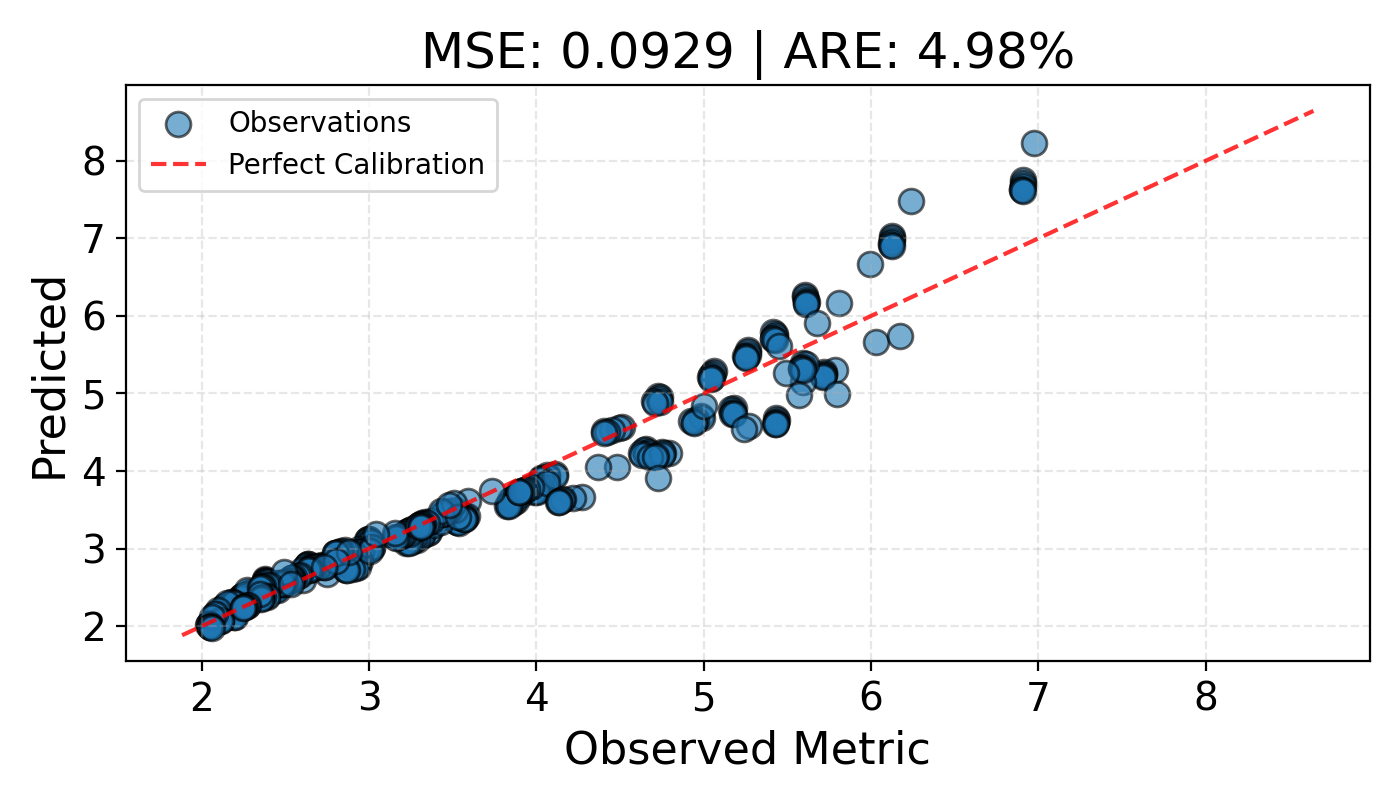}
        \caption{ARC Easy}
        \label{fig:calib_arc_easy}
    \end{subfigure}

    \vskip 0.1cm % Vertical space between rows

    % Row 2
    \begin{subfigure}[b]{0.49\textwidth}
        \centering
        \includegraphics[width=\textwidth]{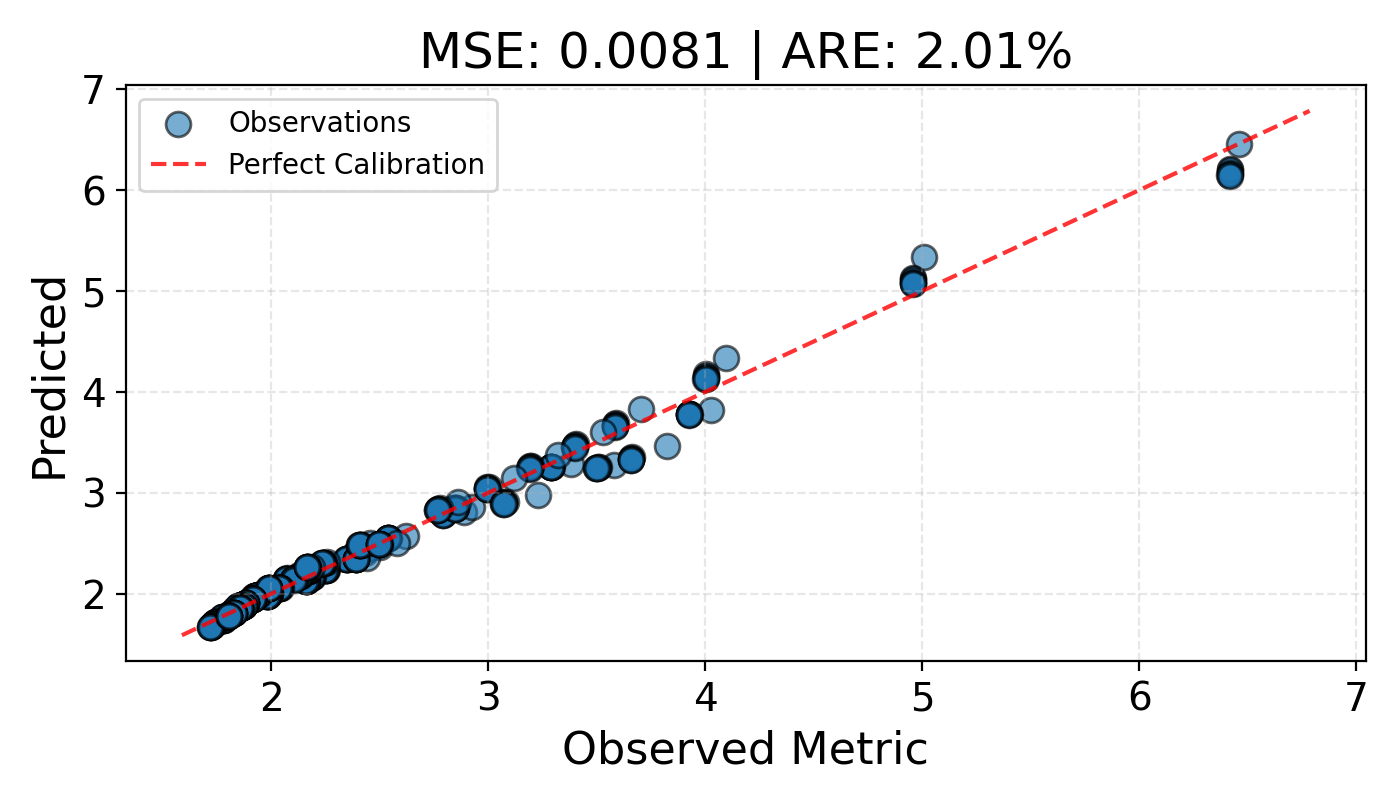}
        \caption{HellaSwag}
        \label{fig:calib_hellaswag}
    \end{subfigure}
    \hfill
    \begin{subfigure}[b]{0.49\textwidth}
        \centering
        \includegraphics[width=\textwidth]{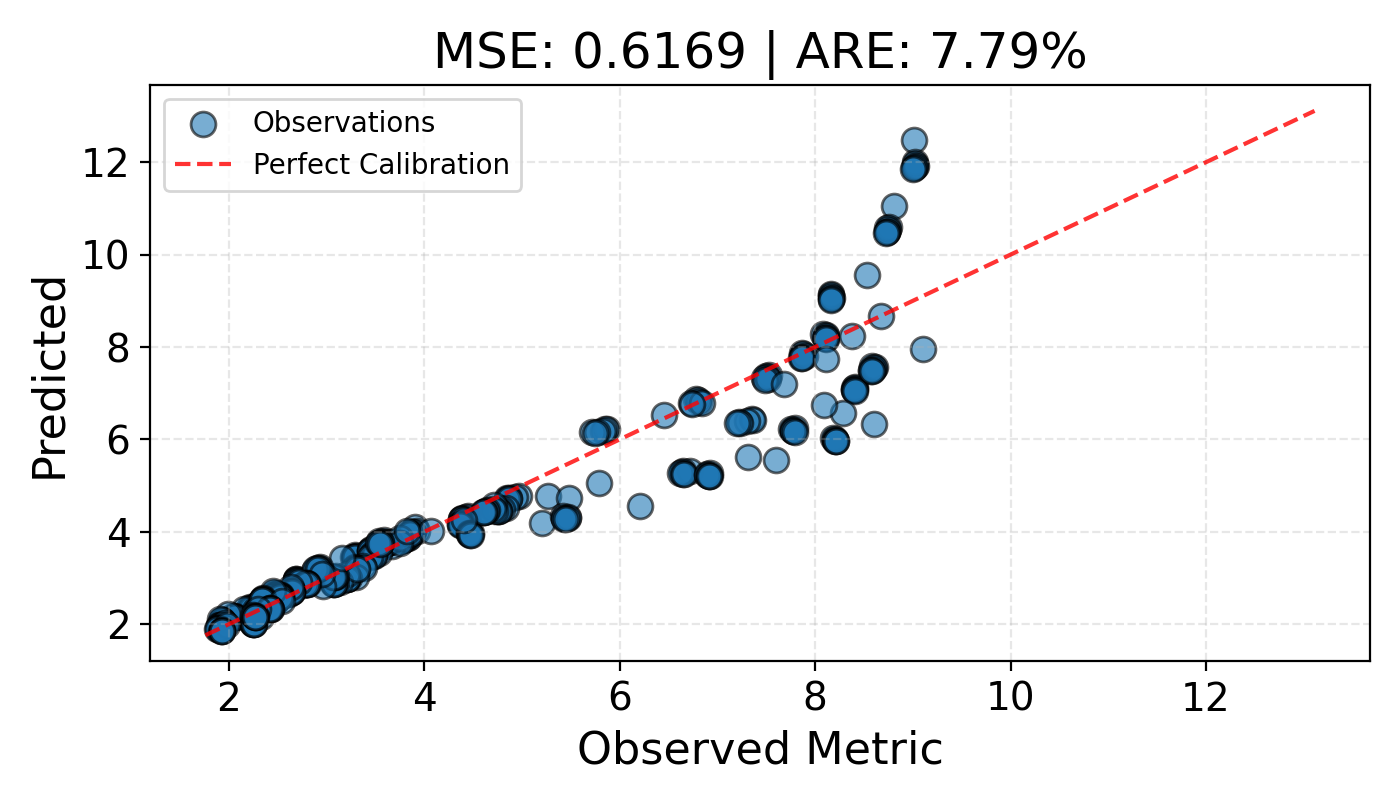}
        \caption{Science Questions (SciQ)}
        \label{fig:calib_sciq}
    \end{subfigure}

    \vskip 0.1cm % Vertical space between rows

    % Row 2
    \begin{subfigure}[b]{0.49\textwidth}
        \centering
        \includegraphics[width=\textwidth]{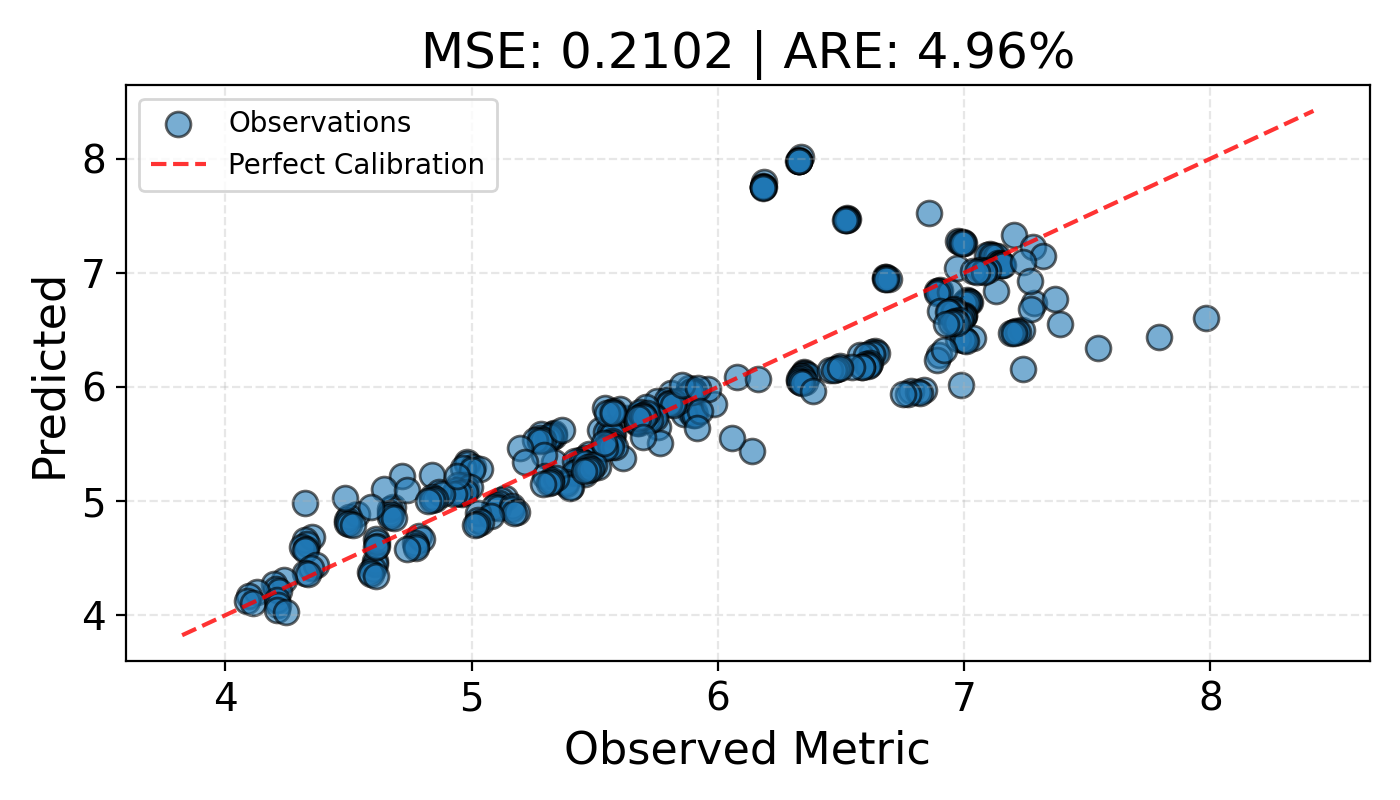}
        \caption{CommonsenseQA}
        \label{fig:calib_commonsenseqa}
    \end{subfigure}
    \hfill
    \begin{subfigure}[b]{0.49\textwidth}
        \centering
        \includegraphics[width=\textwidth]{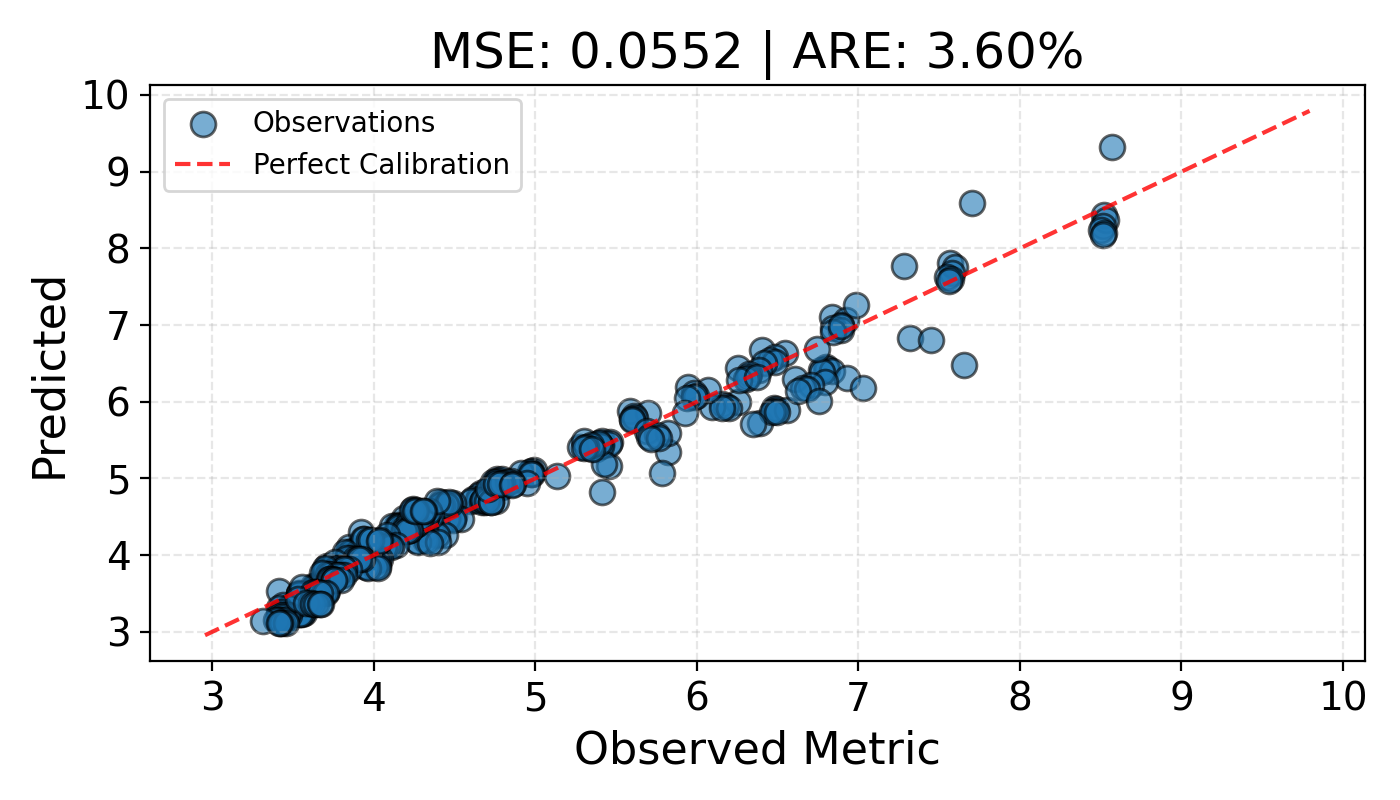}
        \caption{OpenBookQA}
        \label{fig:calib_openbookqa}
    \end{subfigure}
    \caption{\textbf{Calibration plots for 3D scaling law fits across benchmarks.} We show the alignment between predicted and observed gold-answer perplexity across six benchmarks for the interaction model $\mathcal{M}_{NS}$ (Eq.~\ref{eq:3d_scaling_law}), which augments the Hoffmann power-law formulation with a capacity- and saturation-modulated logarithmic retrieval term. The tight grouping around the diagonal indicates that the model effectively captures retrieval-augmented scaling behavior. Reported MSE and ARE values are computed in-sample over the plotted points; held-out cross-validation errors are reported separately in Table~\ref{tab:rag_3d_scaling_fits}.}
    \label{fig:calibration_appendix_group1}
\end{figure*}

Calibration plots (Figure \ref{fig:calibration_appendix_group1}) allow us to visualize how our predicted loss (from the 3D scaling law fit) compares to the actual loss values. 

\subsection{Stability Analysis}
\label{sec:appendix-stability}

To assess the robustness of our scaling-law fits, we evaluate stability across multiple random seeds and model initializations. Specifically, we consider three random seeds each for three model families (30M, 136M, and 233M), yielding a total of 27 runs. For each configuration, we fit scaling laws independently and compute both cross-validation average relative error (CV ARE) and leave-one-model-out ARE (LOMO ARE). Figure~\ref{fig:seed_stability} reports the mean and standard deviation of these metrics across runs for each benchmark. Overall, we observe low variance in both CV ARE and LOMO ARE across most tasks, indicating that the fitted scaling relationships are stable with respect to initialization and data ordering. Reasoning-heavy tasks such as StrategyQA exhibit higher variance and larger absolute errors, suggesting that their scaling behavior is noisier and less well captured by simple parametric forms. In contrast, more knowledge-driven benchmarks (e.g., ARC, OpenBookQA) show consistently low variance and strong fit quality across runs.

% \begin{table}[h]
% \centering
% \small
% \begin{tabularx}{\linewidth}{l >{\raggedright\arraybackslash}X >{\raggedright\arraybackslash}X >{\raggedright\arraybackslash}X}
% \toprule
% \textbf{Dataset} & \textbf{Query (abridged)} & \textbf{Query-only retrieved context (abridged)} & \textbf{Assessment} \\
% \midrule
% SimpleQA & "A Survey of London... which other justice presiding?" &
% Pipe Rolls / King's Council / London court-record style passages &
% Domain-aligned; potentially answer-bearing \\

% SimpleQA & "According to Hegel, what are the three Romantic arts?" &
% Aesthetics/philosophy prose on beauty, criticism, Romanticism &
% Topically strong; likely supports extraction \\

% SimpleQA & "1898 astronomical phenomena: how many eclipses?" &
% Eclipse-focused historical/educational astronomy text &
% Relevant background; exact count not always explicit \\
% \midrule
% GSM8K & "15 gallons... 1/4 container... how many pints?" &
% Dictionary-style "gallon" definitions and mixed web text &
% Occasionally useful unit fact; mostly redundant/noisy \\

% GSM8K & "20% of 50 people... horse #12" &
% Unrelated school snippets, forum text, non-English fragments &
% Mostly distractive; not solution-bearing \\

% GSM8K & "Senior gifts: frame +20%, pins, cords" &
% E-commerce listings/coupons/product pages with prices &
% Lexical overlap only; weak mathematical utility \\
% \bottomrule
% \end{tabularx}
% \caption{Representative query-only retrieval behavior from SimpleQA and GSM8K.}
% \label{tab:qual_query_only_examples}
% \end{table}

\begin{figure*}[t]
\centering %begin{center}
\centerline{\includegraphics[width=\textwidth]{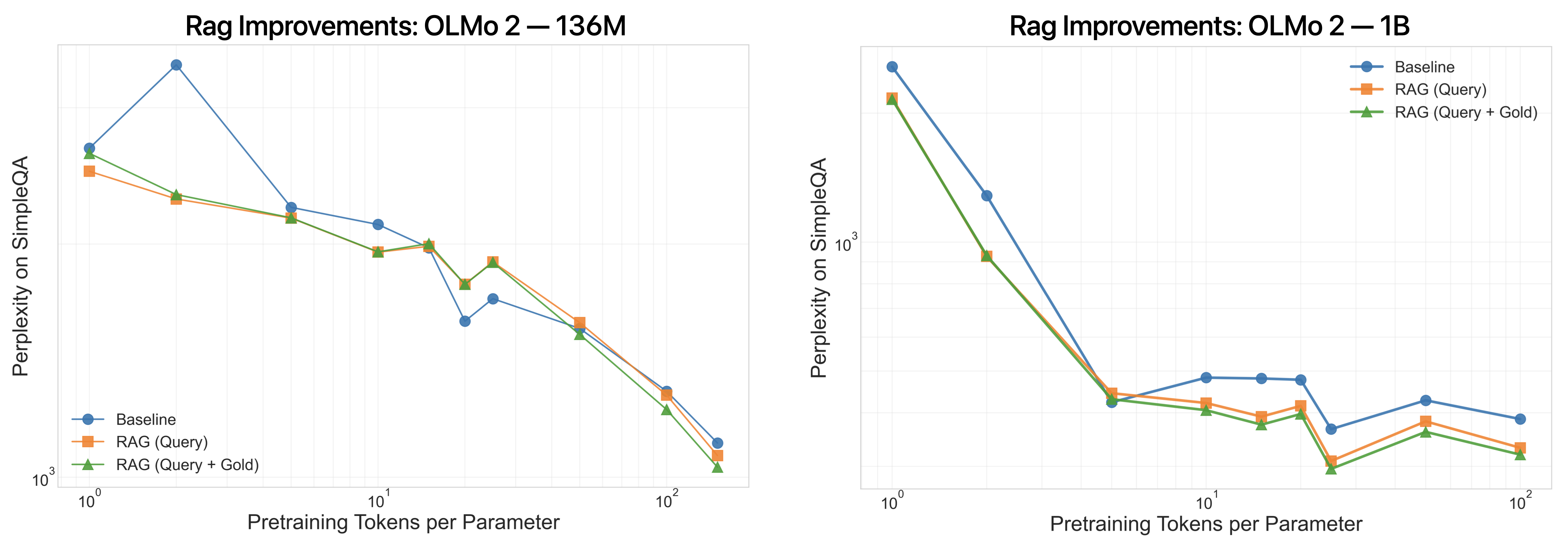}}
% \vskip 0.2cm
\caption{
\textbf{Effect of retrieval query formulation on performance.}
Comparison of standard generation on SimpleQA without retrieval (Baseline) to RAG under two query formulations: (i) \emph{RAG (Query)}, which retrieves top-$k$ passages using only the question, and (ii) \emph{RAG (Query + Gold)}, which includes the gold answer in the query too (an oracle-style ablation). SimpleQA is not multiple-choice (no answer choices), so we do not report \emph{RAG (Query + Choices)} here. All methods use a shared corpus index constructed from 20\% of the data, retrieving the top-5 passages per query. 
\textit{Left:} OLMo-2 136M. \textit{Right:} OLMo-2 1B.
}
\label{fig:rag-improvements}
%\end{center}
% \vskip -0.3in
\end{figure*}

\begin{figure*}[t]
\begin{center}
\centerline{\includegraphics[width=\textwidth]{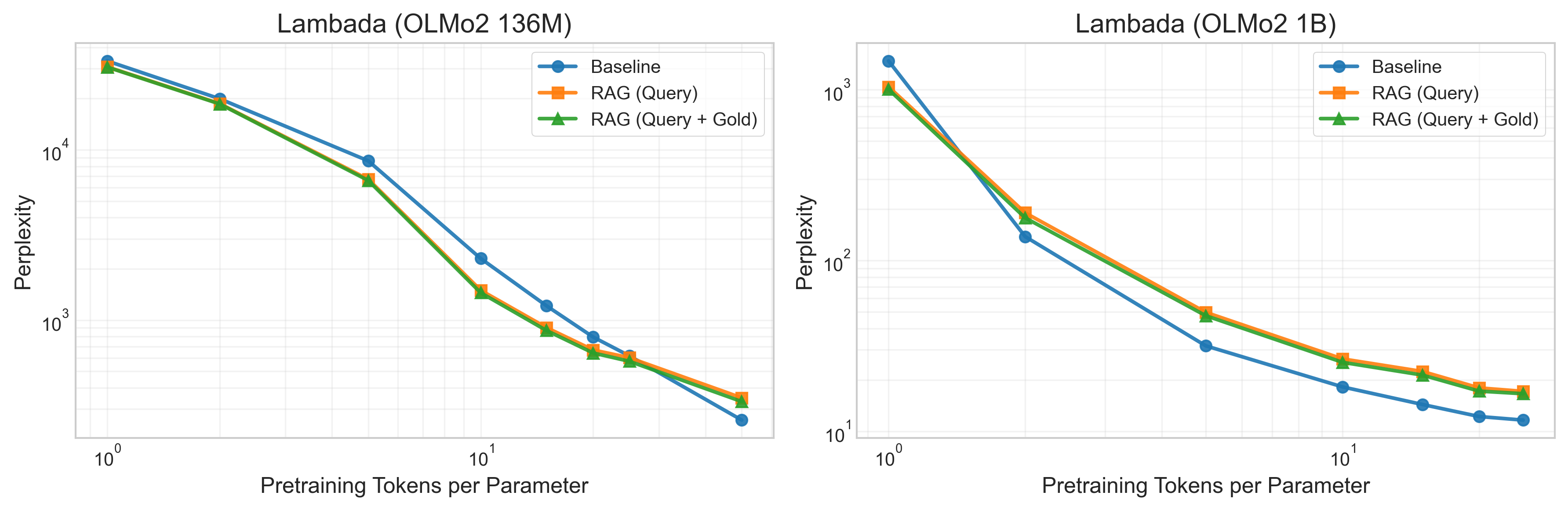}}
\vskip 0.2cm
\caption{
\textbf{Effect of retrieval query formulation on LAMBADA.} Similar to Figure \ref{fig:rag-improvements} above.
}
\label{fig:rag-improvements-lambada}
\end{center}
\vskip -0.2in
\end{figure*}

\begin{figure*}[t]
\begin{center}
\centerline{\includegraphics[width=\textwidth]{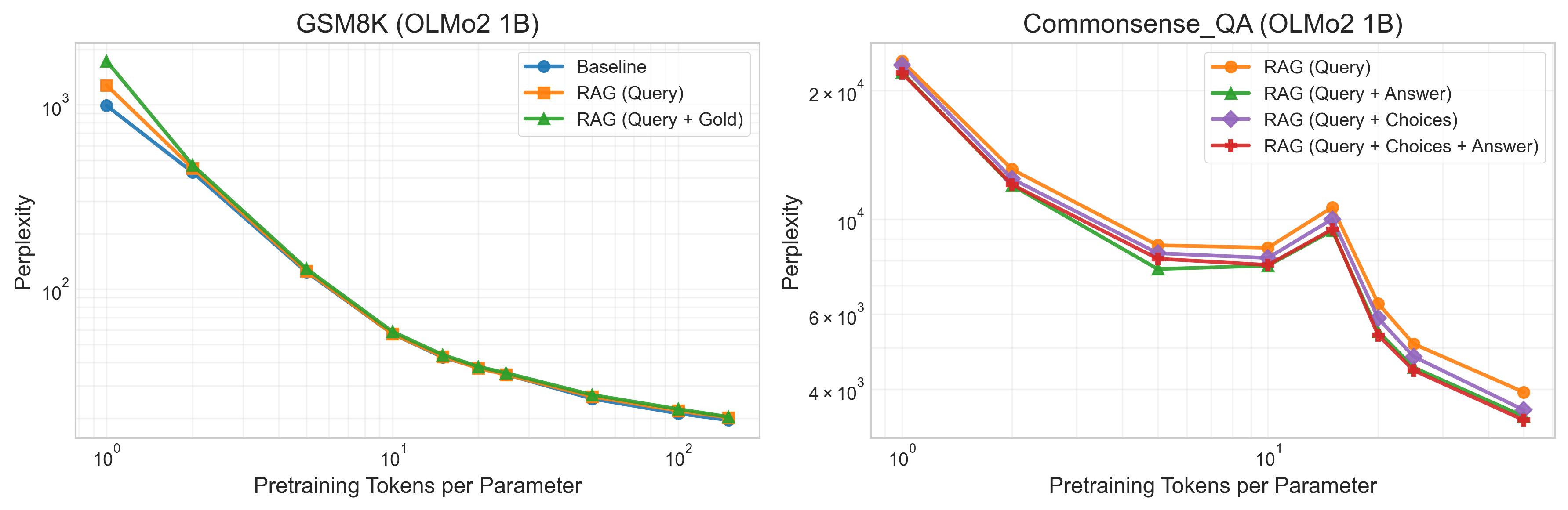}}
\vskip 0.2cm
\caption{
\textbf{Effect of retrieval query formulation across GSM8K and CommonsenseQA.}
Similar to Figure~\ref{fig:rag-improvements}. For CommonsenseQA, we additionally compare querying both choices and gold answer (RAG Query + Choices + Answer).
Both panels show OLMo-2 1B as a function of pretraining tokens per parameter.
}
\label{fig:rag-improvements-gsm8k}
\end{center}
\vskip -0.2in
\end{figure*}

\subsection{RAG Improvements}
\label{sec:appendix-rag-improvements}

While our primary focus is on the allocation trade-off between pretraining and retrieval, this raises a complementary question: how much of the observed benefit depends on retrieval quality? Qualitatively, retrieved contexts on factoid QA often capture the correct topic but do not consistently contain directly answer-bearing evidence (e.g., specific entities or dates), suggesting that retrieval precision remains a limiting factor. To probe this further, we evaluate a simple strategy for improving retrieval: varying retriever query formulation. We compare, with a fixed corpus index across methods, (i) question-only queries, (ii) queries augmented with answer choices (when applicable), and (iii) queries that include the gold answer (an oracle).\footnote{This is closer to an approximate upper bound by more closely approximating an optimal scenario where one would have a (near) \textit{perfect} retriever to maximize the potential benefits of RAG.}

% While our primary focus is on the allocation tradeoff between pretraining and retrieval, this raises a complementary question: how much of the observed benefit depends on retrieval quality itself? Qualitatively, top-$k$ contexts on factoid QA often land in the correct topical neighborhood and provide useful background, but do not always contain directly answer-bearing evidence (e.g., precise names, dates, or ranks) in the highest-ranked passages. This pattern is consistent with a high-recall retrieval stage whose precision can still be limiting for exact-answer tasks (see Appendix \ref{sec:appendix-qual-retrieval} for additional qualitative analysis of retrieval).

All RAG conditions use the same FAISS index, constructed from a held-out corpus whose size is equivalent to 20\% of the maximum pretraining-token budget. We prepend the top-$k=5$ retrieved passages to the prompt and sweep pretraining tokens per parameter while keeping the retrieval configuration fixed.

Figure~\ref{fig:rag-improvements}, \ref{fig:rag-improvements-lambada}, and \ref{fig:rag-improvements-gsm8k} report results on SimpleQA, LAMBADA GSM8K, and CommonsenseQA. Retrieval yields modest gains on knowledge-heavy tasks (SimpleQA, CommonsenseQA), particularly when queries better align with the answer, with improvements increasing at larger model scales. In contrast, reasoning-heavy tasks (GSM8K \citep{cobbe2021trainingverifierssolvemath}, LAMBADA \citep{paperno-etal-2016-lambada}\footnote{We try RAG improvements on these two additional benchmarks (math and word prediction) that performed poorly with retrieval on our initial pilot studies.}), show minimal change. Across tasks, improved retrieval yields incremental gains but does not alter the scaling trends observed earlier. This reinforces prior takeaways that retrieval is not a uniform substitute for pretraining, and that its effectiveness depends on both model scale and task type.

% We provide results analyzing the effect of retrieval query formulation across additional benchmarks. We compare standard generation without retrieval (Baseline) to retrieval-augmented setups using different query constructions. Specifically, we consider: (i) \emph{RAG (Query)}, which retrieves using only the task question; (ii) \emph{RAG (Query + Gold / Answer)}, which augments the query with the gold answer (oracle-style ablation); and for multiple-choice settings, (iii) \emph{RAG (Query + Choices)} when multiple-choice references are available and (iv) \emph{RAG (Query + Choices + Answer)}. All experiments use a shared retrieval setup with a fixed FAISS index constructed from a held-out corpus with size equivalent to 20\% of the max pretraining tokens, and top-$k=5$ retrieved passages prepended to the prompt. We sweep pretraining tokens per parameter while keeping the retrieval configuration fixed. Figures~\ref{fig:rag-improvements-gsm8k} and \ref{fig:rag-improvements-lambada} show results on GSM8K, CommonsenseQA, and LAMBADA across model scales.

\subsection{Qualitative Analysis of Retrieval Behavior}
\label{sec:appendix-qual-retrieval}

To better understand when retrieval helps, we manually inspected retrieved contexts for a couple of datasets where RAG helps (SimpleQA, NQ-Open) and one where it does not (GSM8K). The key contrast is that retrieval is often useful as \emph{topical grounding} for SimpleQA and NQ-Open, but is much less useful for GSM8K, where problems are typically self-contained and external text is often distractive.

For SimpleQA, query-only contexts often match the right domain (e.g., history, philosophy, astronomy, biology) and sometimes surface source-adjacent material that can support answer extraction. For example, legal-history questions about \emph{A Survey of London} retrieve medieval court/tower records; philosophy questions about Hegel retrieve prose that discusses his views on Romantic art and its main art forms; and technical biomedical questions retrieve mutation/gene-focused research text rather than generic web chatter.

For NQ-Open, RAG helps mostly when retrieval injects a single concrete anchor (year/number/entity). For example, \textit{“when was the first Australian prime minister elected”} changes from baseline \textit{1977} to correct \textit{1901} with RAG, and \textit{“what age do you need to be to buy a BB gun”} shifts from \textit{5} to \textit{18 years old} after retrieval includes age-threshold text (\textit{“over the age of 18… over 21 for handguns”}). Likewise, \textit{“when was the last time anyone was on the moon”} moves from \textit{“200 years ago”} to \textit{1972} once Apollo timeline snippets appear. Sometimes the same mechanism still misses the target: retrieval changes the model’s answer, but the new answer remains incorrect. Overall, the benefit is best characterized as occasional fact anchoring from salient cues, rather than consistently reliable evidence grounding.

For GSM8K, query-only retrieval is usually not needed and often noisy. Retrieved passages are commonly worksheets, forum posts, product listings, dictionary pages, or malformed index fragments (e.g., \textit{``files\_...''}), which rarely contribute to the arithmetic decomposition required by the question. Occasionally, retrieval provides a useful conversion fact (e.g., \textit{gallon-to-pints}), but most examples are either redundant with what is already stated in the prompt or off-task. This qualitative pattern aligns with the quantitative result that RAG yields little benefit on GSM8K in this setup, and with existing conclusions that RAG helps more heavily with long-tail factual knowledge than things like mathematics.

\end{document}